\documentclass[runningheads]{llncs}
\usepackage{graphicx}
\usepackage{tikz}
\usepackage{comment}
\usepackage{amsmath,amssymb} % define this before the line numbering.
\usepackage{color}

\usepackage[accsupp]{axessibility}  % Improves PDF readability for those with disabilities.

\usepackage{amsmath} % assumes amsmath package installed
\usepackage{amssymb}  % assumes amsmath package installed
\usepackage{kotex}
\usepackage{comment}
\usepackage{xcolor}
\usepackage{graphicx}
\usepackage{subcaption}
\usepackage{pgfplots}
\usepackage{pifont}
\usepackage{booktabs}
\usepackage{color,soul}
\usepackage{multirow}
\usepackage{wrapfig}
\usepackage{adjustbox}
\usepackage{tabularx}
\usepackage{cite}
\usepackage{mathrsfs}
\usepackage{enumitem}
\usepackage{ulem}

\def\eg{e.g.} 
\def\ie{i.e.} 
\def\ie{i.e.}

\newcommand{\Tref}[1]{Table~\ref{#1}}
\newcommand{\Eref}[1]{Eq.~(\ref{#1})}
\newcommand{\Fref}[1]{Fig.~\ref{#1}}
\newcommand{\Sref}[1]{Section~\ref{#1}}

\usepackage{xcolor}
\definecolor{myRed}{RGB}{200, 36, 37}
\definecolor{myGreen}{RGB}{42,152,42}
\definecolor{myGray}{RGB}{122, 122, 122}
\definecolor{myYellow}{RGB}{188, 189, 34}
\definecolor{myBlue}{RGB}{24, 96, 145}

\begin{document}

\pagestyle{headings}

\mainmatter

\title{SLiDE: Self-supervised LiDAR De-snowing through Reconstruction Difficulty}

\titlerunning{SLiDE: Self-supervised LiDAR De-snowing through Reconstruction Difficulty}

\author{Gwangtak Bae\inst{1} \and
Byungjun Kim\inst{1} \and
Seongyong Ahn\inst{2} \and\\
Jihong Min\inst{2} \and
Inwook Shim\inst{3}\thanks{Inwook Shim is the corresponding author.}}

\authorrunning{G. Bae et al.}

\institute{\textsuperscript{1}Seoul National University, \textsuperscript{2}Agency for Defense Development, \textsuperscript{3}Inha University \email{\{tak3452,peterbj95,seongyong.ahn,happymin77\}@gmail.com, iwshim@inha.ac.kr}}

\maketitle

\begin{abstract}
LiDAR is widely used to capture accurate 3D outdoor scene structures. However, LiDAR produces many undesirable noise points in snowy weather, which hamper analyzing meaningful 3D scene structures. Semantic segmentation with snow labels would be a straightforward solution for removing them, but it requires laborious point-wise annotation. To address this problem, we propose a novel self-supervised learning framework for snow points removal in LiDAR point clouds. Our method exploits the structural characteristic of the noise points: low spatial correlation with their neighbors. Our method consists of two deep neural networks: Point Reconstruction Network (PR-Net) reconstructs each point from its neighbors; Reconstruction Difficulty Network (RD-Net) predicts point-wise difficulty of the reconstruction by PR-Net, which we call \textit{reconstruction difficulty}. With simple post-processing, our method effectively detects snow points without any label. Our method achieves the state-of-the-art performance among label-free approaches and is comparable to the fully-supervised method. Moreover, we demonstrate that our method can be exploited as a pretext task to improve label-efficiency of supervised training of de-snowing.
\keywords{LiDAR de-snowing, vision for adverse weather, self-supervised de-snowing}
\end{abstract}
\section{Introduction}

Robust and accurate 3D scene measurement is an essential component of outdoor machine perceptions, \eg, autonomous vehicles. LiDAR is a commonly used 3D measurement sensor that gives reliable 3D point clouds in favorable weather conditions. However, in snowy weather conditions, LiDAR frequently generates a large number of particle noise points by detecting solid snowflakes~\cite{pitropov2020canadian}. These noise points could have a fatal impact on point cloud applications for outdoor systems~\cite{kilic2021lidar,gruber2019pixel,shim2016vision}.

Conventional filter-based approaches~\cite{charron2018noising, park2020fast, rusu20113d} have been presented for the LiDAR de-noising task to alleviate this problem. They attempt to remove the noise points by evaluating their spatial vicinity, but these approaches often suffer from misclassification since they only rely on simple spatial sparsity. Following the success of deep learning in various 3D point cloud applications~(\eg, classification~\cite{qi2017pointnet, qi2017pointnet++}, detection~\cite{zhou2018voxelnet, shi2019pointrcnn}, segmentation~\cite{landrieu2018large, milioto2019rangenet++}, etc.), a deep learning-based LiDAR de-noising approach, WeatherNet~\cite{heinzler2020cnn}, is recently introduced. It takes advantage of deep learning-based semantic segmentation methods to detect point-wise LiDAR noise points.

\begin{figure}[t!]
    \centering
    \begin{subfigure}{0.33\linewidth}
    \includegraphics[width=\linewidth]{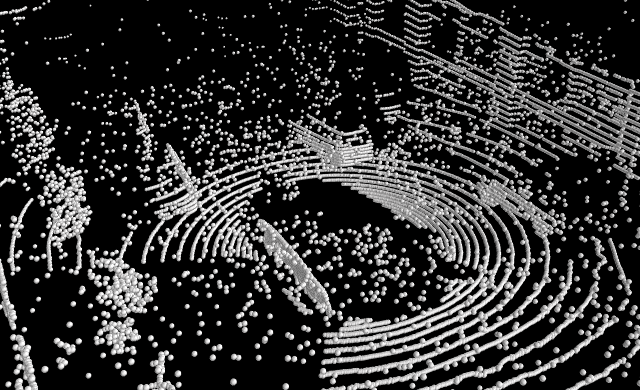}
    \caption{Noisy point cloud}
    \label{first:noisy}
    \end{subfigure}\hfill
    \begin{subfigure}{0.33\linewidth}
    \includegraphics[width=\linewidth]{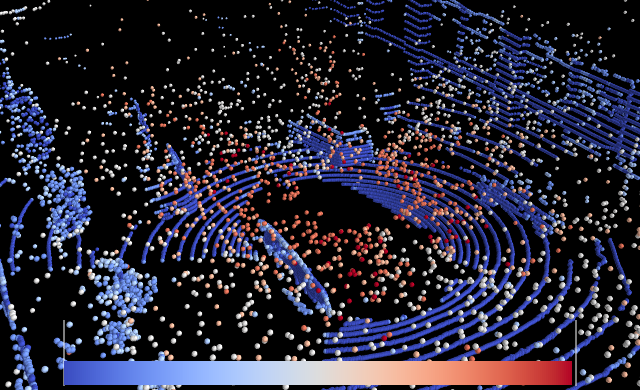}
    \caption{Reconstruction difficulty}
    \label{first:cmap}
    \end{subfigure}\hfill
    \begin{subfigure}{0.33\linewidth}
    \includegraphics[width=\linewidth]{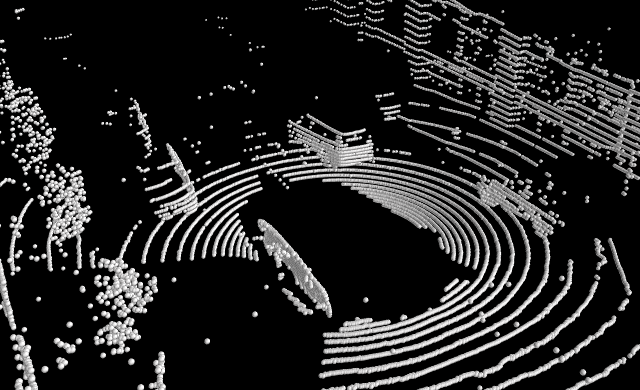}
    \caption{De-snowed point cloud}
    \label{first:denoised}
    \end{subfigure}\hfill
    \caption{An example of the proposed LiDAR de-snowing process estimating how difficult to reconstruct each point from neighboring points. In (b), the closer to the red, the more difficult to reconstruct.
    }
    \label{fig_first}
\end{figure}

While WeatherNet~\cite{heinzler2020cnn} outperforms the conventional approaches by a significant margin, it requires point-wise annotations. Even though there were attempts on efficient 3D point annotation~\cite{luo2015patch, luo2018semantic, piewak2018boosting}, manual annotation of 3D point cloud is still laborious and time-consuming~\cite{gao2021we}. Moreover, labeling snow noise points requires even more efforts. Labeling the snow points is hard to take advantage of the assistance of camera view, which is the convention in point cloud labeling process~\cite{geiger2012we, caesar2020nuscenes, sun2020scalability}, since camera images are also heavily degraded, \eg, snowflakes on the lens and in the air. Besides, in most cases, the degradation is not consistent with the noise points of LiDAR.

To address this issue, we propose a self-supervised learning framework for de-noising LiDAR point clouds in snowy weather, \ie, \textit{LiDAR de-snowing}. We focus on the structural characteristic of the noise points in snowy weather that they have low spatial correlations with their neighbors. Therefore, they are difficult to be reconstructed from their neighboring points. 
Based on this insight, we propose a novel self-supervised learning approach for snow points removal in LiDAR point clouds. Our method consists of two deep neural networks: Point Reconstruction Network~(PR-Net) reconstructs randomly selected target points from their neighbors; Reconstruction Difficulty Network~(RD-Net) predicts point-wise error of PR-Net reconstruction, which we call \textit{reconstruction difficulty}. An example of the estimated reconstruction difficulty and the de-snowing result is visualized in~\Fref{fig_first}. A set of reconstruction target points for training is selected from the point cloud itself, which leads to a self-supervised training scheme. The two deep neural networks are jointly trained with a shared loss function, and then only RD-Net, followed by simple post-processing, is used to detect the noise points in the inference step. Our model trained without any labeled data outperforms previous label-free approaches with a large margin and achieves a comparable performance to the fully supervised model.

While our self-supervised method is basically label-free, it can be extended to a semi-supervised training scheme where training data consist of a small amount of labeled data and a large amount of unlabeled data. We demonstrate that an extension of our method as a pretext task enables supervised training of de-snowing in a label-efficient manner. Our semi-supervised extension yields better performance with fewer labeled data, which shows that our method as a pretext task is well-aligned with supervised training of de-snowing.

Our contributions can be summarized as follows:
\begin{itemize}[leftmargin=*]
\item To the best of our knowledge, we are the first to introduce a self-supervised approach for LiDAR de-noising under snowy weather, \ie, \textit{LiDAR de-snowing}. Our method identifies noise points that have low spatial correlations with their neighboring points.
\item Our method outperforms previous label-free approaches by a large margin and achieves comparable results to the supervised method.
\item We demonstrate that our self-supervised approach can be exploited as a pretext task for supervised de-snowing, which largely improves label-efficiency.

\end{itemize}
\section{Related Work}

\subsection{LiDAR Point Cloud De-noising}

Conventional filter-based methods classify noise points by their spatial sparsity~\cite{rusu20113d, charron2018noising, park2020fast, roriz2021dior, zhou2019dup}.
Radius outlier removal~(ROR)~\cite{rusu20113d} finds the number of neighbors within a fixed radius and classifies points as noises if the number of neighbors is lower than a predefined threshold. ROR often fails at far-distant points because the density of the LiDAR point cloud decreases as the distance increases. Dynamic radius outlier removal~(DROR)~\cite{charron2018noising} employs varying search radii according to the distance to overcome the limitation of the previous works and successfully removes snow noise points. While those filter-based methods are easy to use and have a low computational burden, they still cannot handle point cloud's irregularity well, which often results in a significantly worse performance.

Recently, a supervised learning-based approach, WeatherNet~\cite{heinzler2020cnn}, is introduced. It formulates the LiDAR de-noising problem as 2D semantic segmentation on the range image representation. Despite WeatherNet outperforms the filter-based methods, it requires expensive point-wise annotated training data.

\subsection{Self-supervised Image De-noising}
Self-supervised deep learning approaches have accomplished remarkable advances in the image de-noising task. Noise2Noise~\cite{lehtinen2018noise2noise} is a pioneering work that restores a noisy image using another noisy image generated from the same clean image source.
Since it is difficult to obtain a pair of noisy images in dynamic scenes, following works~\cite{krull2019noise2void, batson2019noise2self} improved the idea to work with a single noisy image by predicting a de-noised version of each pixel without depending on the pixel itself.

Despite the success of self-supervised de-noising in the image domain~\cite{lehtinen2018noise2noise, krull2019noise2void, batson2019noise2self, quan2020self2self, kim2021noise2score} and dense point cloud domain~\cite{hermosilla2019total, luo2020differentiable, luo2021score}, there are problems in utilizing those methods for the LiDAR de-noising task in snowy weather, \ie, the \textit{de-snowing} task. First, the image de-noising cannot produce sufficiently reliable results when a pixel value is difficult to be precisely predicted by its neighboring pixel information~\cite{krull2019noise2void}. Second, the objective of LiDAR de-noising~\cite{rusu20113d, charron2018noising, park2020fast, roriz2021dior} mainly focuses on removing noise points, not restoring the original points. For many applications that use LiDAR data~(\eg, grasping an object, avoiding obstacles), rather than restoring the clean version of every elements, which is the main objective of the image de-noising tasks, it is essential to discard unreliable measurements while preserving clean measurements.

\subsection{Semi-supervised Learning}
Semi-supervised learning is a training scheme to learn from both labeled data and unlabeled data~\cite{van2020survey}. How to design an unsupervised loss function for leveraging sufficient unlabeled data is the main concern of semi-supervised learning. The categories of semi-supervised learning methods include unsupervised pretraining followed by fine-tuning~\cite{jarrett2009best, le2013building, devlin2018bert, xie2020pointcontrast}, consistency regularization~\cite{sajjadi2016regularization, laine2016temporal, tarvainen2017mean}, pseudo labeling~\cite{lee2013pseudo, iscen2019label, xie2020self}, and combination of these methods~\cite{berthelot2019mixmatch, sohn2020fixmatch, zoph2020rethinking}. While our method is label-free, \ie, self-supervised, we demonstrate that our method can be extended to a semi-supervised training scheme. Combining our self-supervised loss with a supervised loss from a limited number of labeled data, the large performance gain shows that our method as a pretext task is well-aligned with supervised learning, which leads to an effective semi-supervised training scheme.
\section{Proposed Method}
\Sref{sec_input} describes the representation of input LiDAR data. \Sref{sec_train} explains our self-supervised learning framework for detecting noise points. Section~\ref{sec_mhl} and \ref{sec_post} give detailed information on multi-hypothesis point reconstruction and post-processing process, respectively. Section~\ref{sec_semi} explains the extension of our self-supervised method into a semi-supervised training scheme.

\subsection{Input Representation} \label{sec_input}
The input to the proposed method is a $2$D range image representation, which is the raw data structure of rotating LiDAR~\cite{fan2021rangedet, milioto2019rangenet++, meyer2019lasernet}, \eg, Velodyne LiDAR. Such representation simplifies the $3$D position reconstruction problem into $1$D depth reconstruction along the LiDAR rays. The range image is generated as follows: Let $P$ be a finite LiDAR point cloud, which contains $K$ points: ${P}=\{\textbf{p}_1, ..., \textbf{p}_K\}$. Each point $\textbf{p}=(p_x, p_y, p_z)$ is projected onto the image plane as $\textbf{u}=(u,v)$ via a mapping function $\mathrm{\Pi}:\mathbb{R}^3\mapsto\mathbb{R}^2$. By following \cite{meyer2019lasernet, heinzler2020cnn}, column $u$ is defined by $u=(\pi - arctan(p_y, p_x))/\delta_h$, where $\delta_h$ is the horizontal resolution of the LiDAR we used. Row $v$ represents the laser id of $\textbf{p}$, which corresponds to one of the sender/receiver modules in LiDAR. The projected point at $(u,v)$ has a range value $r=(p_x^2+p_y^2+p_z^2)^{1/2}$. For each scan of LiDAR, we generate a corresponding range image $\mathrm{R \in \mathbb{R}^{n \times m}}$.

\begin{figure*}[t!]
  \includegraphics[width = 0.95 \linewidth,trim={0 0mm 0 0},clip]{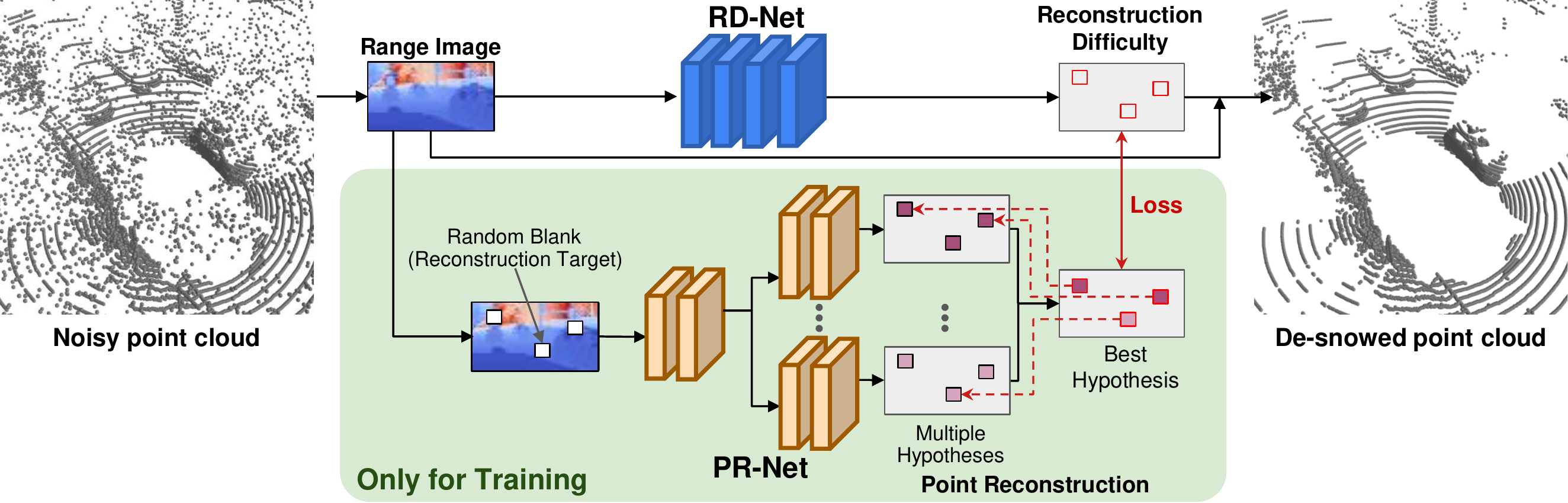}
  \centering
  \caption{The overall structure of our proposed LiDAR de-snowing method.}
  \label{fig_highlevel}
\end{figure*}

\begin{comment}
\begin{figure}[t]
  \includegraphics[width = 1.0\columnwidth]{figures/Concept_Figure_MHL.pdf}
  \centering
  \caption{Example of ambiguity inherent in the point reconstruction task. PR-Net with multiple hypotheses can reconstruct both of the multiple plausible answers.
  }
  \label{fig_concept_mhl}
\end{figure}
\end{comment}

\subsection{Self-supervised Learning Framework} \label{sec_train}
We propose a self-supervised approach of LiDAR de-snowing that does not require point-wise annotations. To detect noise points that have low spatial correlations to their neighboring points, we designed a point reconstruction task and utilized errors from the task as guidance for training our de-snowing network. As shown in~\Fref{fig_highlevel}, reconstruction target points are randomly selected in the range image for every iteration of training. The Point Reconstruction Network~(PR-Net) then predicts depth values of the target points by aggregating information of their neighboring points. At the same time, the Reconstruction Difficulty Network~(RD-Net) estimates the extent of errors of the points reconstructed by PR-Net, which we call \textbf{\textit{reconstruction difficulty}}.

\begin{figure}[t]
  \includegraphics[width = \columnwidth]{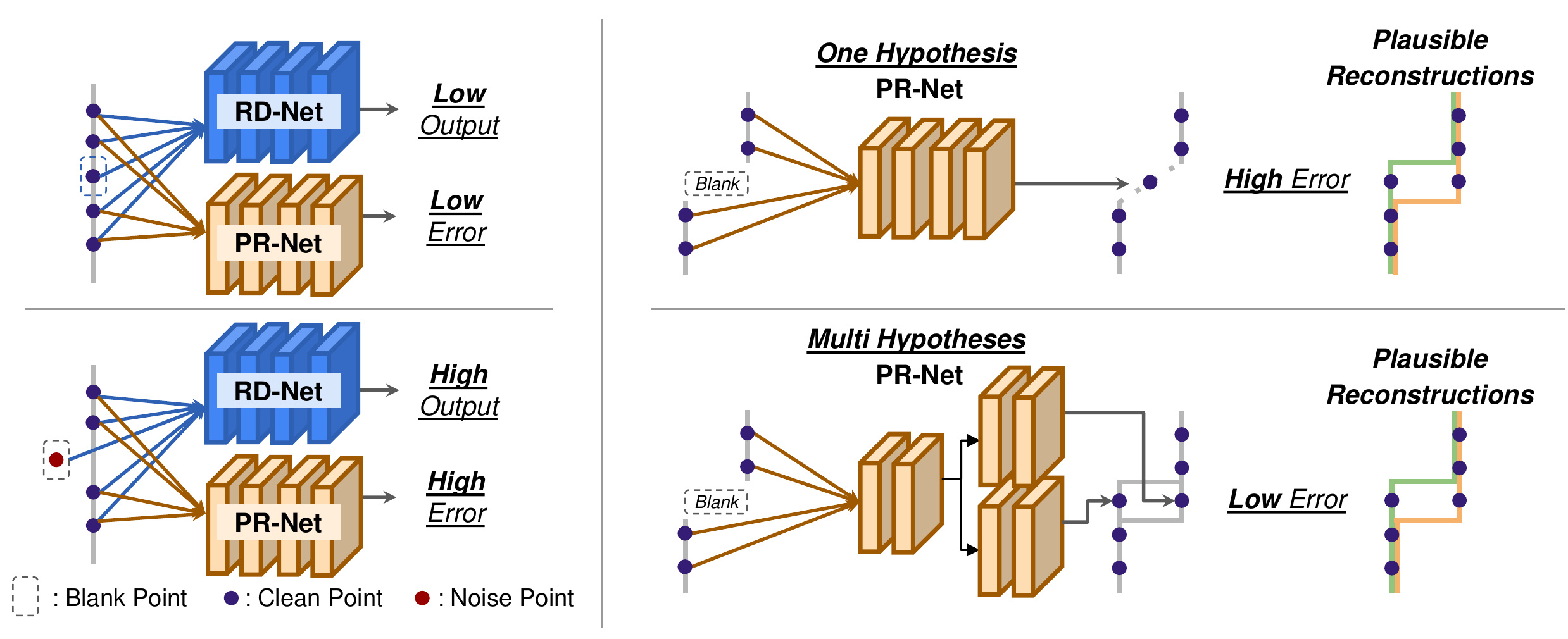}
  \centering
  \caption{The left image explains how PR-Net and RD-Net infer differently depending on whether a clean point or a noise point is blanked. The right image presents an example of ambiguity inherent in the point reconstruction task. PR-Net with multiple hypotheses can infer both of the multiple plausible reconstructions.}
  \label{fig_concept_pr_nd}
\end{figure}

The two deep neural networks, PR-Net and RD-Net, are jointly trained with a shared loss function. We design a loss function with two objectives. First, as seen in the left figure in~\Fref{fig_concept_pr_nd}, the loss function should guide RD-Net to produce high output for noise points on which PR-Net has high error. Second, when training the point reconstruction task, the loss function should attenuate the loss contribution of noise points because they are extremely difficult to be reconstructed. It is to make PR-Net concentrate on reconstructing non-noise points. Our baseline loss function is given as follows:
\begin{equation} \label{eq_loss_baseline}
\begin{split}
    \mathcal{L}_{self} &=\sum I \odot \left[ \sqrt{2}\frac{ \left| \theta(\widetilde{R})-R \right| }{\exp(\phi(R))} + \phi(R) \right], %\\
\end{split}
\end{equation}
where $R$ is a range image of LiDAR data and $\widetilde{R}$ is a randomly blanked range image. $I$ is a binary mask that only selects loss from the blanked points. $\odot$ is an element-wise multiplication. $\theta(\cdot)$ and $\phi(\cdot)$ denote PR-Net and RD-Net, respectively. The blanked points are selected randomly for each iteration of training. The structure of~\Eref{eq_loss_baseline} is inspired by the negative log-likelihood of a Laplacian distribution~\cite{kendall2017uncertainties,klodt2018supervising}.

PR-Net’s reconstruction error guides RD-Net to learn to estimate the reconstruction difficulty of each point. PR-Net takes the blanked range image~$\widetilde{R}$ as an input and predicts the depth value of the blanked points, and then we calculate $L1$ loss. \Eref{eq_loss_baseline} guides~$\exp(\phi(R))$ to be high when the reconstruction error~$L1$ is expected to be high, and to be low for the opposite. In other words, RD-Net is trained to predict the expected reconstruction quality of PR-Net for each point by taking the original range image~$R$, which is not blanked, as an input. Consequently, we can identify the noise points by only utilizing the prediction of RD-Net. The regularization term~$\phi(R)$ is added to prevent RD-Net from predicting an infinite value.

\Eref{eq_loss_baseline} also allows training of PR-Net to be robust to noise points. Trying to minimize the loss on noise points could decrease the reconstruction ability for clean points. Therefore, in~\Eref{eq_loss_baseline}, gradients from noise points are attenuated by $\exp(\phi(R))$ as a weighting factor for the loss given to PR-Net. Points with high output from RD-Net have a smaller effect on the loss of PR-Net.

At the test time, only RD-Net is used to detect noise points. It takes a range image~$R$ as an input and predicts how difficult it will be to reconstruct each point. If the output of RD-Net $\phi(R)$ is higher than a certain threshold, the point is classified as a noise point.

\subsection{Point Reconstruction with Multiple Hypotheses} \label{sec_mhl}
PR-Net is trained to reconstruct randomly selected target points and their reconstruction errors are computed. However, the point reconstruction task has an inherent ambiguity, where some clean points may have multiple plausible answers for reconstruction. For example, in the right of~\Fref{fig_concept_pr_nd}, when a point on the object boundary is selected as a target, there can be two plausible answers: the object boundary or background. Since PR-Net with a single output cannot cover both answers precisely, the output is collapsed to the mean of them, which leads to an undesirable increase of the reconstruction error for clean points. 

In order to distinguish the ambiguous clean points and noise, PR-Net is modified to have multiple outputs as shown in~\Fref{fig_highlevel}, which is inspired by multi-hypothesis learning. Following ~\cite{yang2019inferring, lee2016stochastic}, only predictions with a minimum error are used for loss calculation. Such prediction for each point is given as follows:
\begin{equation} \label{eq_argmin}
\begin{split}
    C_i = \min_k\left| \theta^{k}(\tilde{R})-R \right|_i,
\end{split}
\end{equation}
where $\theta^{k}(\cdot)$ is the $k^{th}$ output, \ie, hypothesis, of PR-Net and $C_{i}$ indicates the output of point $i$ with a minimum error. Then, our reconstruction loss in \Eref{eq_loss_baseline} can be modified as follows:
\begin{equation} \label{eq_loss_mhl}
\begin{split}
    \mathcal{L}_{self, mhl} &=\sum I \odot \left[ \sqrt{2}\frac{C}{\exp (\phi(R)) } + \phi(R) \right].
\end{split}
\end{equation}
The modified loss guides PR-Net to have multiple plausible predictions rather than to be collapsed to the mean. If at least one of the multiple predictions is well-reconstructed, the reconstruction error will be low. However, since noise points are still difficult to reconstruct, albeit with a finite number of multiple predictions, reconstruction errors will be high.

\begin{figure}[t!]
    \centering
    \hspace{11mm}
    \begin{subfigure}{0.35\linewidth}
    \includegraphics[width=\linewidth]{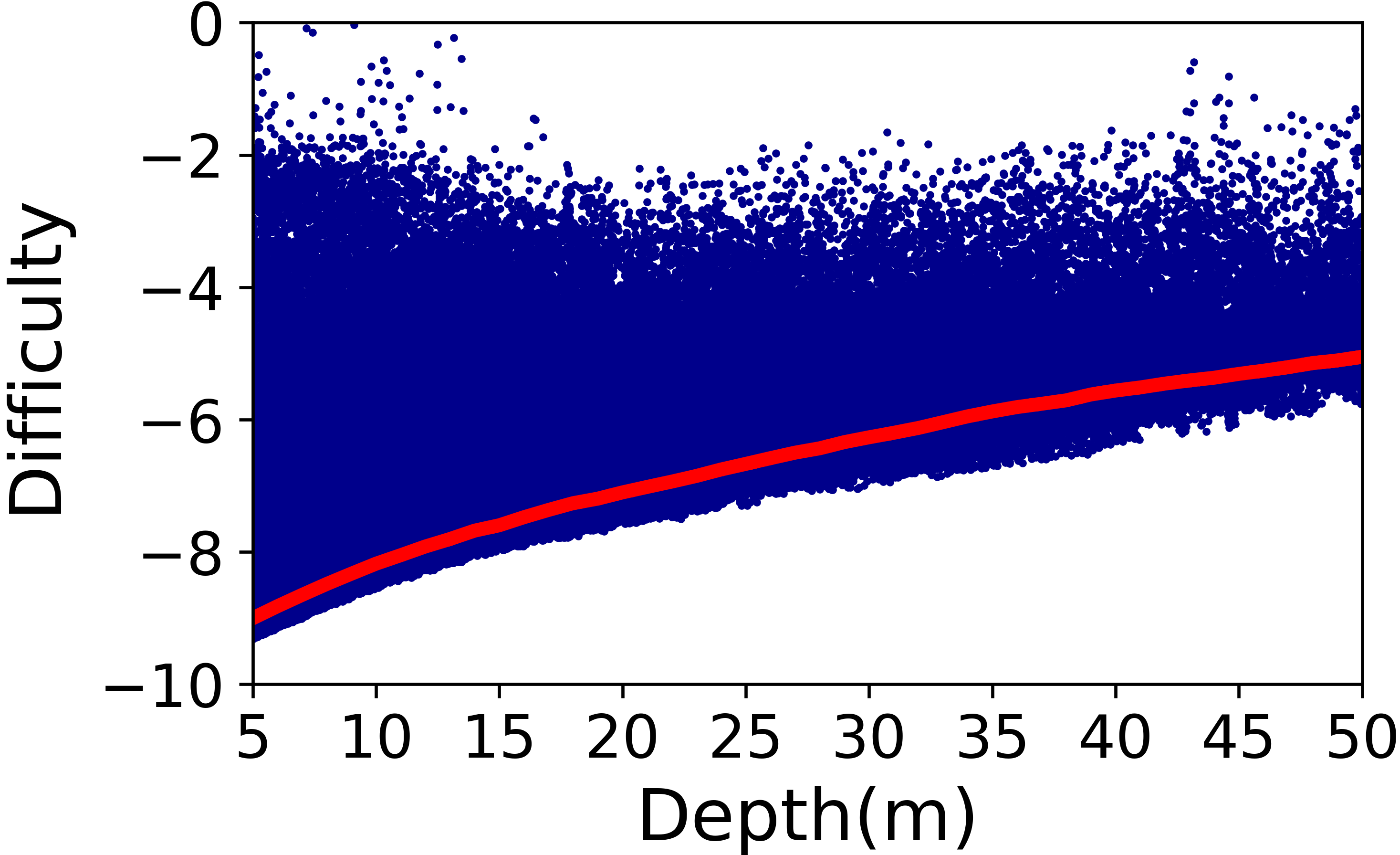}
    \caption{Before post-processing}
    \label{u_shift:before}
    \end{subfigure}\hfill
    \begin{subfigure}{0.35\linewidth}
    \includegraphics[width=\linewidth]{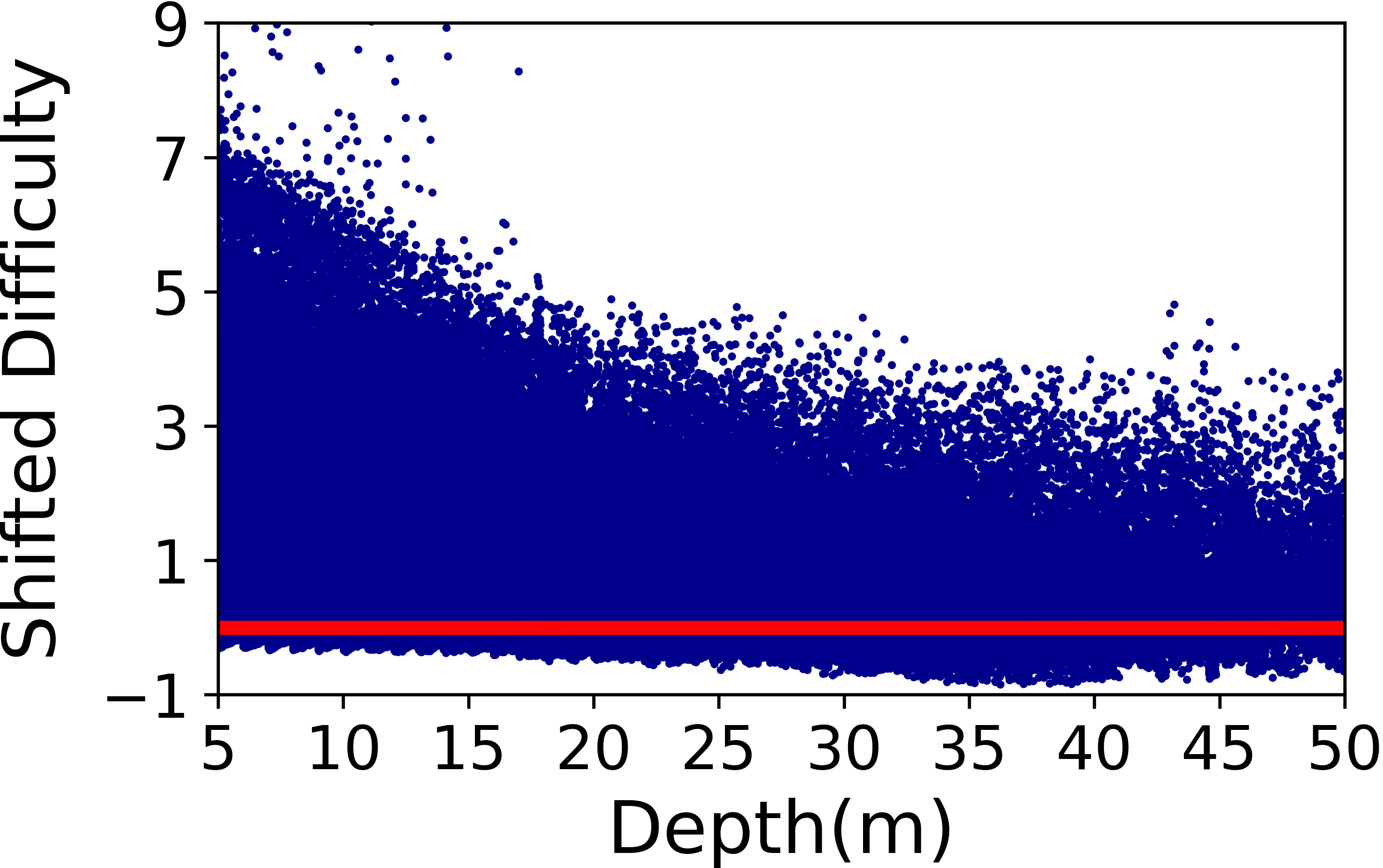}
    \caption{After post-processing}
    \label{u_shift:after}
    \end{subfigure}
    \hspace{11mm}
    \vspace{-2mm}
    \caption{RD-Net's output of each point in order of depth before~(a) and after post-processing~(b). Red lines indicate the $20^{th}$ percentile of each meter of depth.}
    \label{fig_uncertainty_shifting}
\end{figure}

\subsection{Post-processing} \label{sec_post}
As the sensing distance increases, the minimum output of RD-Net gradually increases, as shown in~\Fref{u_shift:before}. This can be seen as the effects of the decreased density of the point cloud at a far distance, which makes the reconstruction more difficult.
To compensate this effect, with partitioning the points by $1m$ depth interval, RD-Net’s output is shifted downward in the amount of the shifting parameter for each depth interval. For example, in~\Fref{u_shift:after}, the shifting parameter is defined as the $20^{th}$ percentile of RD-Net's output value. It compensates the depth-dependent bias on reconstruction difficulty. We then detect the noise points based on a certain threshold that is empirically determined. The ablation studies on the shifting parameter are described in~\Sref{para_study}.

\begin{figure}[t]
  \includegraphics[width =0.9\columnwidth]{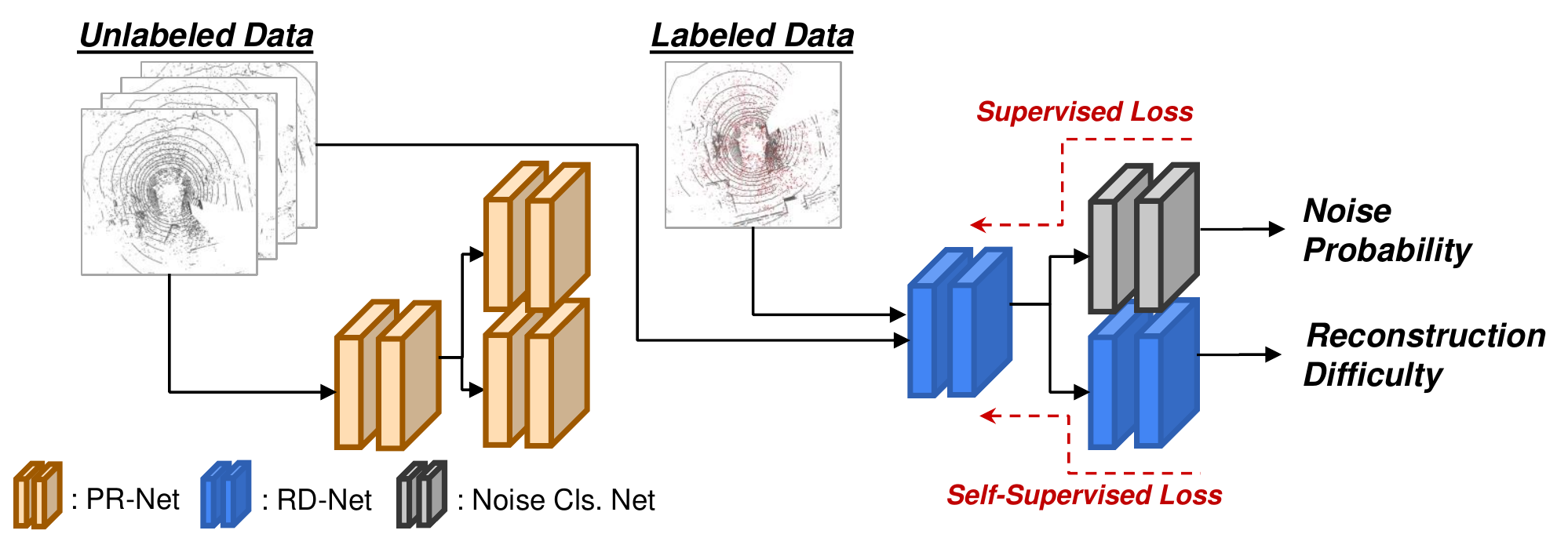}
  \centering
  \caption{Architecture design of our semi-supervised extension. The feature encoder of RD-Net and the noise classification network is shared.}
  \label{fig_concept_ssl}
\end{figure}

\begin{figure}[t!]
    \centering
    \begin{subfigure}{0.33\linewidth}
    \includegraphics[width=\linewidth]{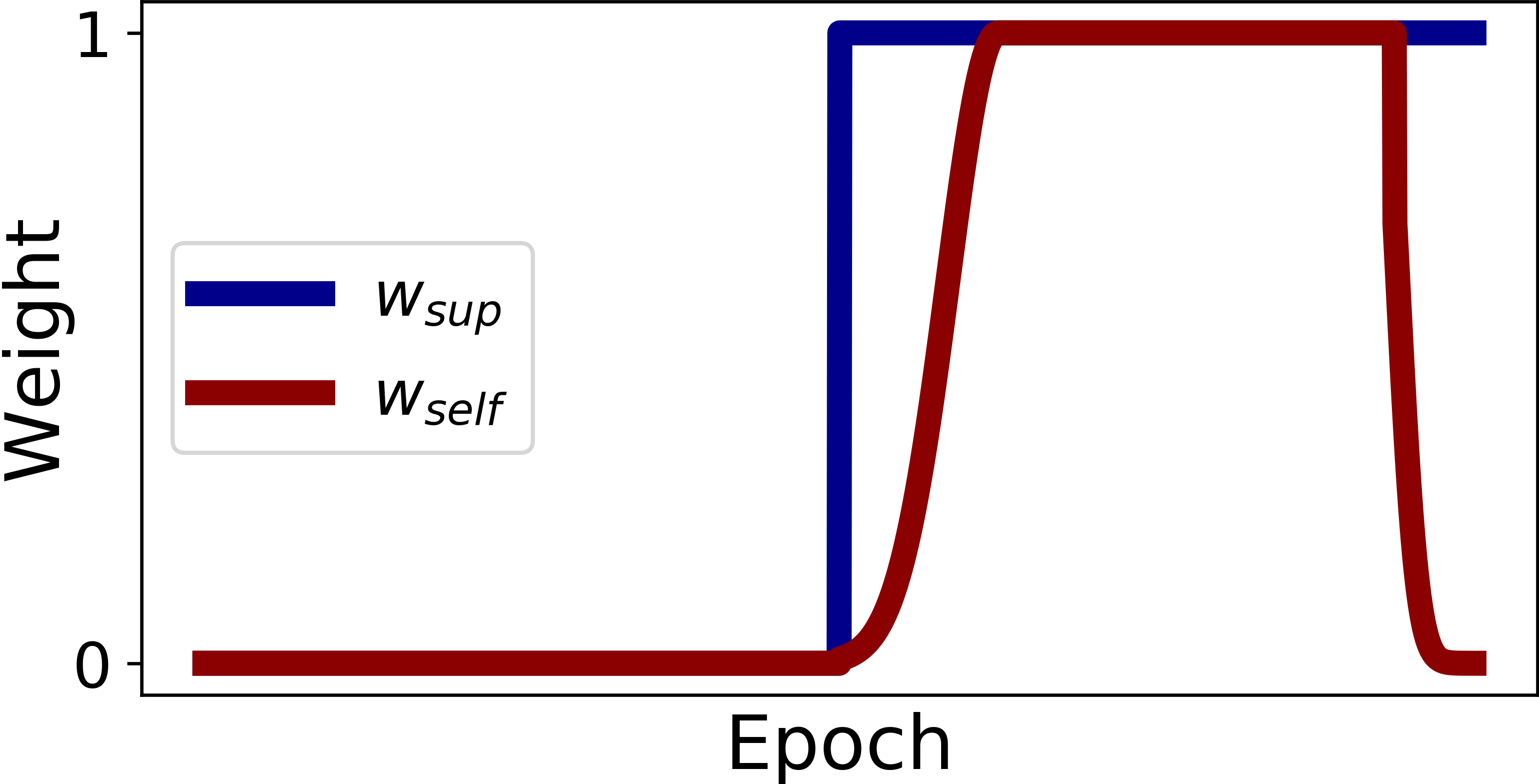}
    \caption{Ramp up/down}
    \label{weight:pi}
    \end{subfigure}\hfill
    \begin{subfigure}{0.33\linewidth}
    \includegraphics[width=\linewidth]{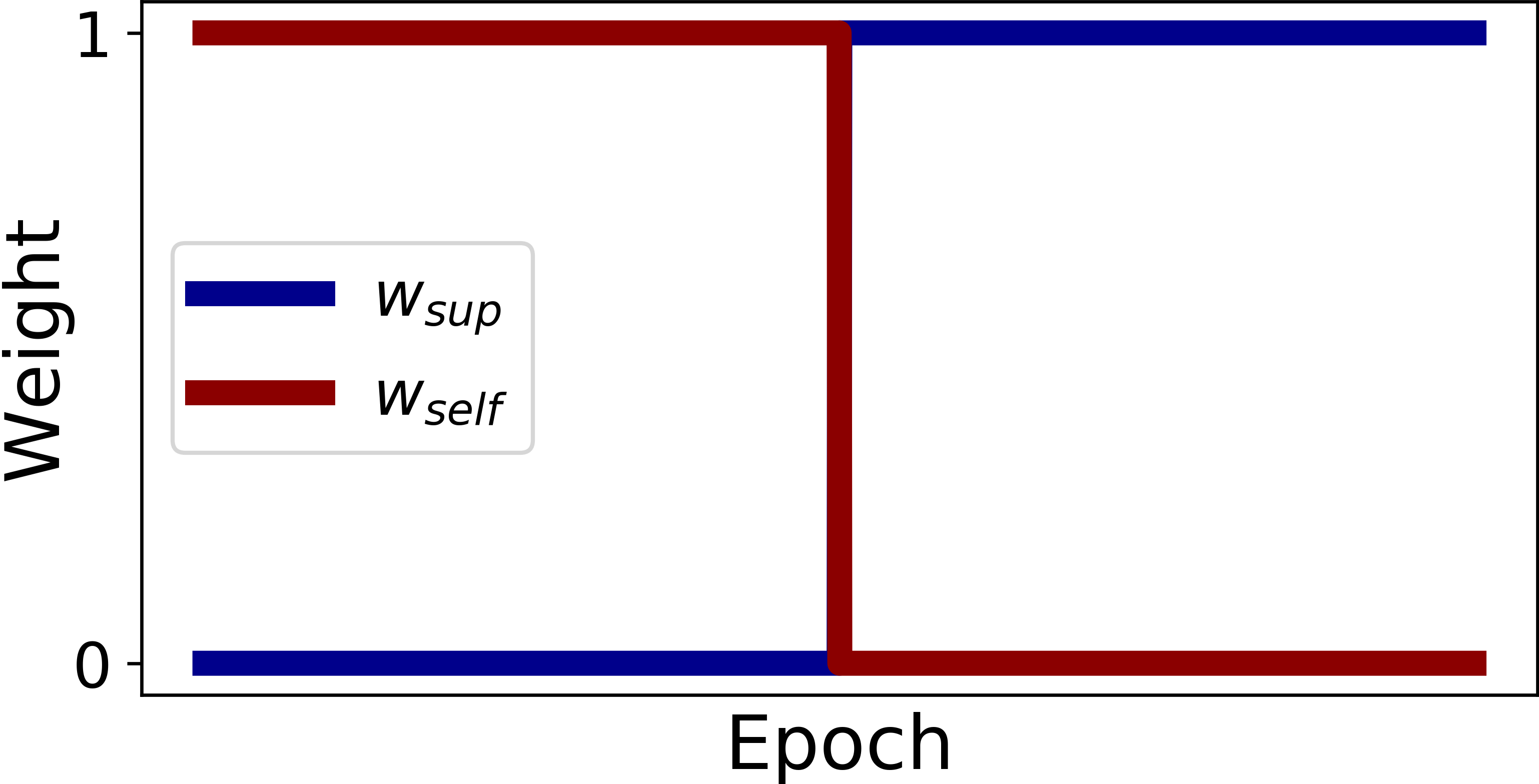}
    \caption{Pretrain}
    \label{weight:pretrain}
    \end{subfigure}\hfill
    \begin{subfigure}{0.33\linewidth}
    \includegraphics[width=\linewidth]{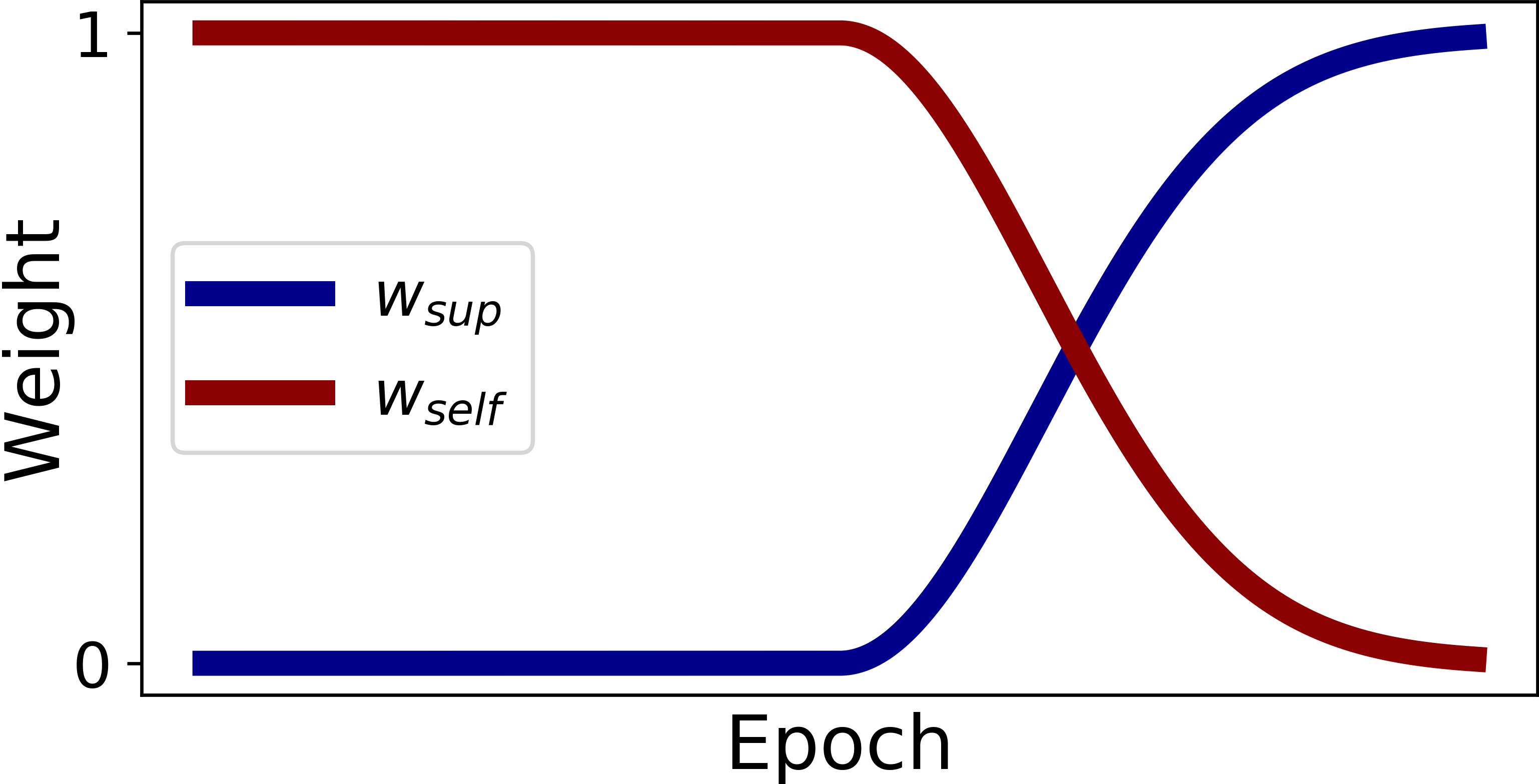}
    \caption{Smooth transfer}
    \label{weight:ours}
    \end{subfigure}\hfill
    \caption{Weighting functions for combining the supervised and self-supervised loss.}
    \label{fig_weight}
\end{figure}

\subsection{Semi-supervised Learning} \label{sec_semi}
While our self-supervised method is able to remove noise points without any labeled data, we found out that it can also be extended to a semi-supervised training scheme. Since RD-Net is trained to regress the reconstruction difficulty, which is one of the characteristics to distinguish noise points from others, we can expect that the backbone network of RD-Net extracts features suitable for noise classification. As shown in~\Fref{fig_concept_ssl}, the feature encoder of RD-Net is shared for the feature encoder of a noise classification network. All of the networks are jointly trained with the weighted sum of two loss functions as follows,
\begin{equation} \label{eq_loss_semi}
\begin{split}
    \mathcal{L} &= w_{self}*\mathcal{L}_{self, mhl} + w_{sup}*\mathcal{L}_{sup},
\end{split}
\end{equation}
where $L_{sup}$ is the cross entropy loss by following WeatherNet~\cite{heinzler2020cnn}. $w_{self}$ and $w_{sup}$ are the weights for the self-supervised loss and the supervised loss, respectively.

Three different weighting functions are proposed and evaluated. \Fref{weight:pi} indicates weighting functions that are widely used in consistency regularization methods for semi-supervised learning~\cite{laine2016temporal, luo2018smooth, tarvainen2017mean, ke2019dual}. Weighting functions in~\Fref{weight:pretrain} are inspired by self-supervised learning~\cite{doersch2015unsupervised, noroozi2016unsupervised, gidaris2018unsupervised, grill2020bootstrap, chen2021exploring}. Our self-supervised method is regarded as a pretext task. The feature encoder of the noise classification network is initialized with the feature encoder of RD-Net. We further extend \Fref{weight:pretrain} into \Fref{weight:ours}. The pretrained features for estimating reconstruction difficulty are smoothly transferred into the noise classification network. The equations for the weighting functions are explained in the supplementary material.
\section{Experimental Result}
\subsection{Point-wise Evaluation on Synthetic Data} \label{result_synthetic}
\subsubsection{Dataset}
A number of scene data with point-wise annotations should be provided for a fair quantitative evaluation, regardless of whether training is done in a supervised or an unsupervised manner. However, to the best of our knowledge, there is no published dataset that satisfies it. We collect real-snow noise points using our stationary data-capturing system and synthesize the collected noise points into various clean weather road scenes.

Since a background point cannot be detected if a noise point is on the same LiDAR ray, we can directly pick out noises by comparing range images of noisy point cloud sets~(\textit{Noise-Set}) and clean point cloud sets~(\textit{Clean-Set}) as follows:
\begin{equation} \label{noise_acquisition}
\begin{split}
    L^{N}_{(u,v)} \leftarrow
    \begin{cases}
        \text{\textbf{N}~(Noise)}, & \text{if } R^{F}_{(u,v)} \geq R^N_{(u,v)}+\tau\\
        \text{\textbf{C}~(Clean)}, & \text{else}
    \end{cases}
\end{split}
\end{equation}
where $R^F$ is the reference range image generated from \textit{Clean-Set}, $R^N$ is a range image of~\textit{Noise-Set}, $L^N$ indicates a label map of $R^N$ and $\tau$ is a margin for sensing errors of LiDAR. In the case of scanning an empty space~(\eg, sky), since no valid background information is projected to $R^{F}_{(u,v)}$, we always assign the `Noise' label to $L_{(u,v)}^{N}$.
The collected noise points are then injected into road scene point cloud sets (\textit{Base-Set}) taken in clean weather as follows:
\begin{equation} \label{noise_injection}
\begin{split}
    R^S_{(u,v)} = 
    \begin{cases}
        R^{N}_{(u,v)}, & \text{if } R^{N}_{(u,v)} \leq R^{max}_{(u,v)}\text{ and }L^{N}_{(u,v)}=\text{\textbf{N}~(Noise)}
        \\
        R^B_{(u,v)}, & \text{else}
    \end{cases}
\end{split}
\end{equation}
\begin{table*}[t] \centering
    \centering
    \newcolumntype{A}{ >{\centering\arraybackslash} m{2.6cm} }
    \newcolumntype{B}{ >{\centering\arraybackslash} m{2.6cm} }
    \newcolumntype{C}{ >{\centering\arraybackslash} m{1.8cm} }
    \caption{Quantitative results on the synthesized snow noise data. \textit{Labeled data} indicates the number of labeled training data used.}
    \begin{tabular}{A|B|C C C}
        \toprule
        Method & Labeled Data & IoU & Precision & Recall\\ \midrule
        ROR~\cite{rusu20113d} & 0 & 17.82 & 17.86 & 98.65\\
        DROR~\cite{charron2018noising} & 0 & 33.68 & 33.87 & 98.37\\
        Ours w/o MHL & 0 & 65.75 & 70.38 & 90.89\\
        Ours & 0 & 79.62 & 85.69 & 91.83\\ \midrule
        WeatherNet~\cite{heinzler2020cnn} & 239  & 41.40 & 76.78 & 47.32 \\
        Ours(Semi.sup.)  & 239  & 82.44 & 96.39 & 85.07 \\ \midrule
        WeatherNet~\cite{heinzler2020cnn} & 23,908  & 84.04 & 97.48 & 85.90\\
        Ours(Semi.sup.)  & 23,908  & 84.24 & 97.24 & 86.30\\
        \bottomrule
    \end{tabular}
    \label{table_noise}
\end{table*}
where $R^B$ is a range image in \textit{Base-Set},
and $R^S$ is a synthesized range image. $R^{max}$ is the maximum detectable range which is introduced for a realistic synthesis~\cite{heinzler2020cnn, bijelic2020seeing} by considering scene structures of $R^B$
as follows:
\begin{equation} \label{detectable_range}
\begin{split}
    R^{max}_{(u,v)} = \min(\frac{-ln(\frac{n}{I^B_{(u,v)} + g})}{2*\beta}, R^{B}_{(u,v)}),
\end{split}
\end{equation}
given the received laser intensity $I^B_{(u,v)}$, the adaptive laser gain $g$, the atmospheric extinction coefficient $\beta$, and the detectable noise floor $n$. In this paper, Nuscenes dataset~\cite{caesar2020nuscenes} is used as \textit{Base-Set} to synthesize snowy scenes. Please see the supplementary material for more details of the dataset generation process.

\subsubsection{Implementation Details}
Our self-supervised model is trained with the synthesized snow noise dataset. We split a total of $34,139$ scans into training, validation, and test sets at a ratio of $70:15:15$. 
PR-Net and RD-Net consist of residual blocks~\cite{he2016deep} for the self-supervised model. For the semi-supervised model, PR-Net and RD-Net use the same backbone with WeatherNet to ensure a fair comparison. Separated layers in~\Fref{fig_concept_ssl} consist of a \textit{LiLaBlock}~\cite{piewak2018boosting} and a convolution layer by following WeatherNet. Each training step encounters an independently selected random set of target points in order to learn various cases of the reconstruction.
Horizontal flipping is randomly performed to augment training data. 
At the test time, points with an RD-Net output higher than the threshold are classified as noise points. Throughout experiments in this paper, the threshold is determined as $2.9$, which achieves the highest performance for the validation dataset.
Quantitative performance is evaluated using the Intersection-over-Union~(IoU) metric, following WeatherNet~\cite{heinzler2020cnn}.

\begin{figure*}[t!]
    \centering
    \begin{subfigure}{0.32\linewidth}
    \includegraphics[width=\linewidth]{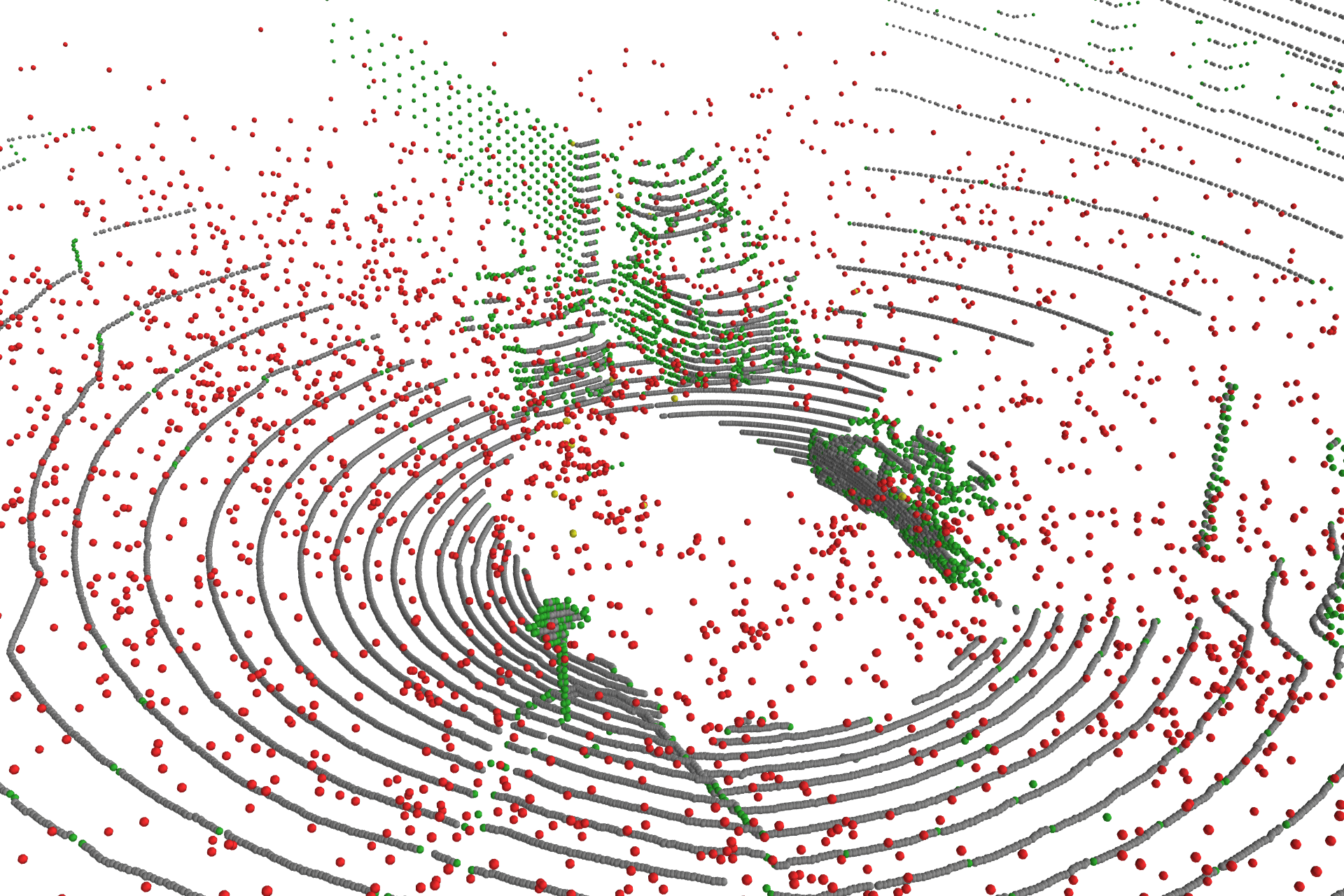}
    \label{fig_result_32:dror1}
    \end{subfigure}\hfill
    \begin{subfigure}{0.32\linewidth}
    \includegraphics[width=\linewidth]{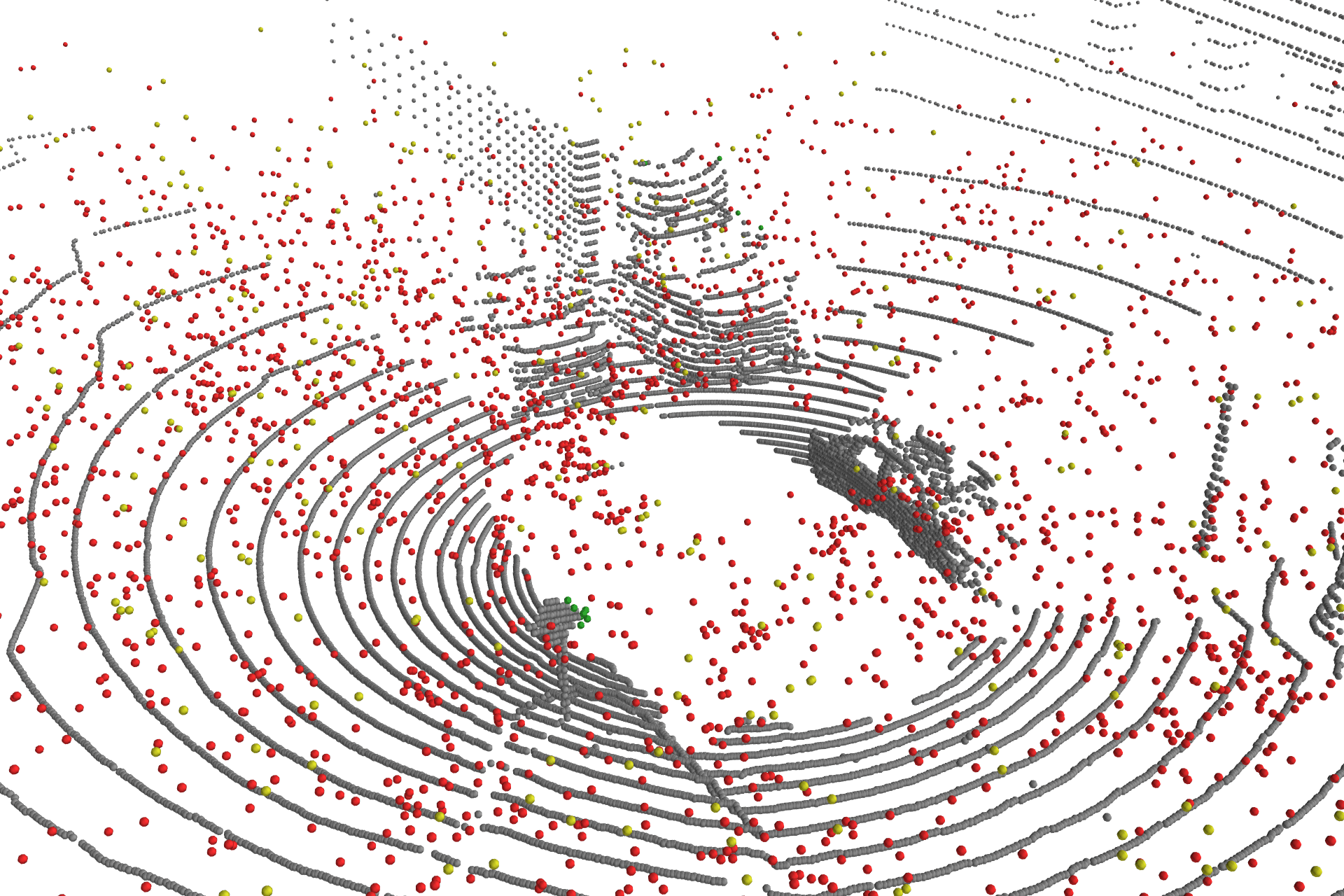}
    \label{fig_result_32:weathernet1}
    \end{subfigure}\hfill
    \begin{subfigure}{0.32\linewidth}
    \includegraphics[width=\linewidth]{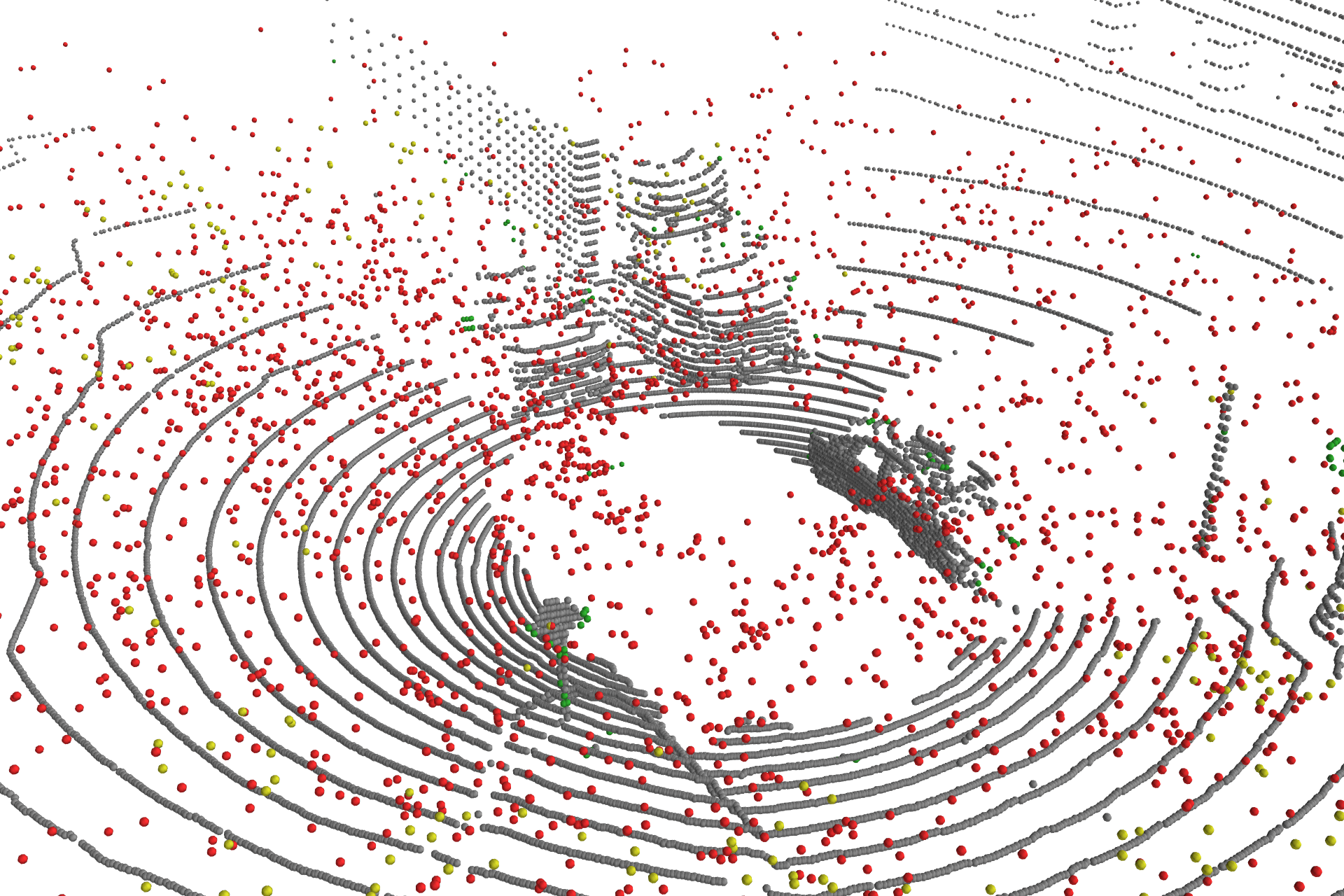}
    \label{fig_result_32:ours1}
    \end{subfigure}\hfill
    \begin{subfigure}{0.32\linewidth}
    \includegraphics[width=\linewidth]{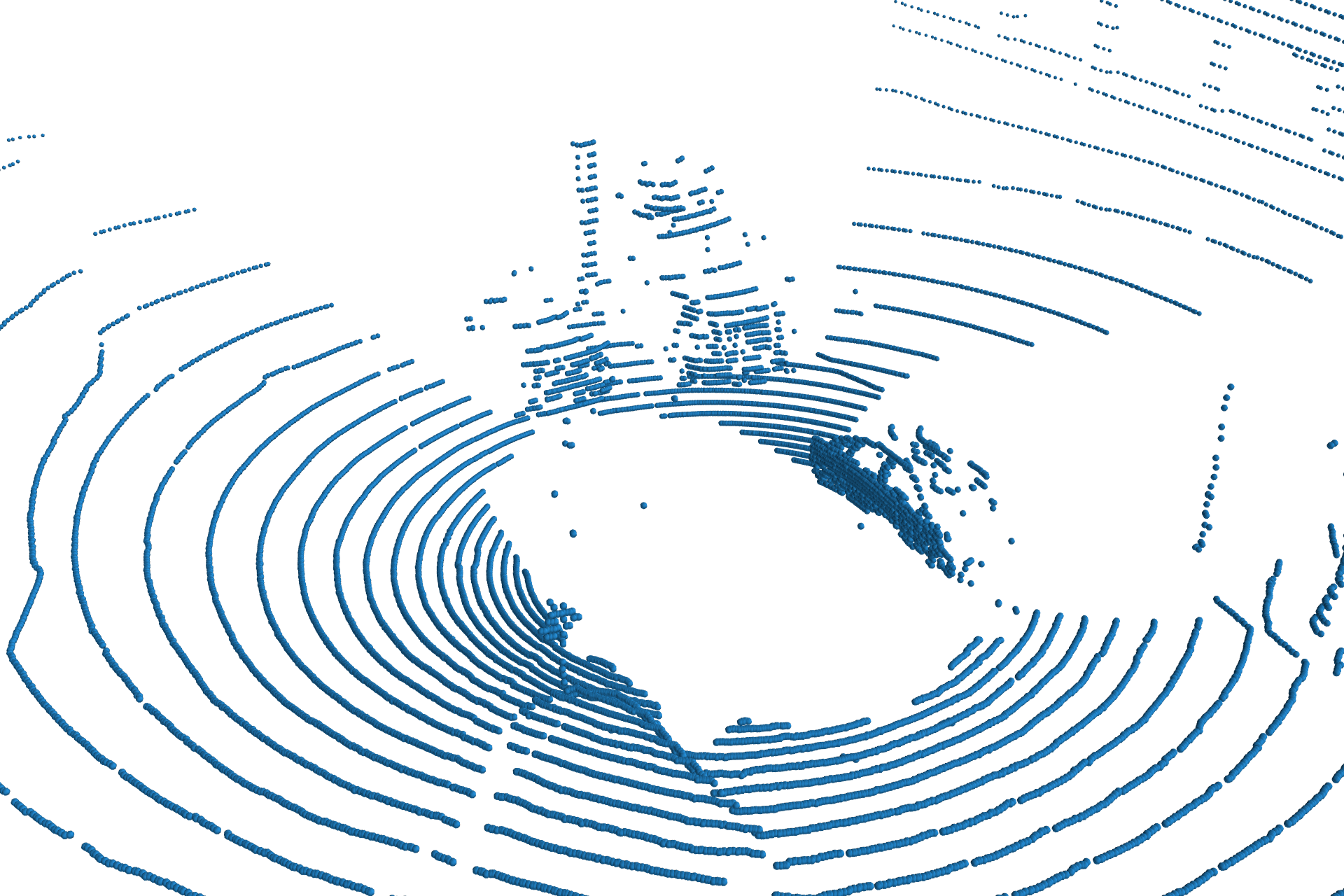}
    \caption{DROR~\cite{charron2018noising}}
    \label{fig_result_32:dror1_d}
    \end{subfigure}\hfill
    \begin{subfigure}{0.32\linewidth}
    \includegraphics[width=\linewidth]{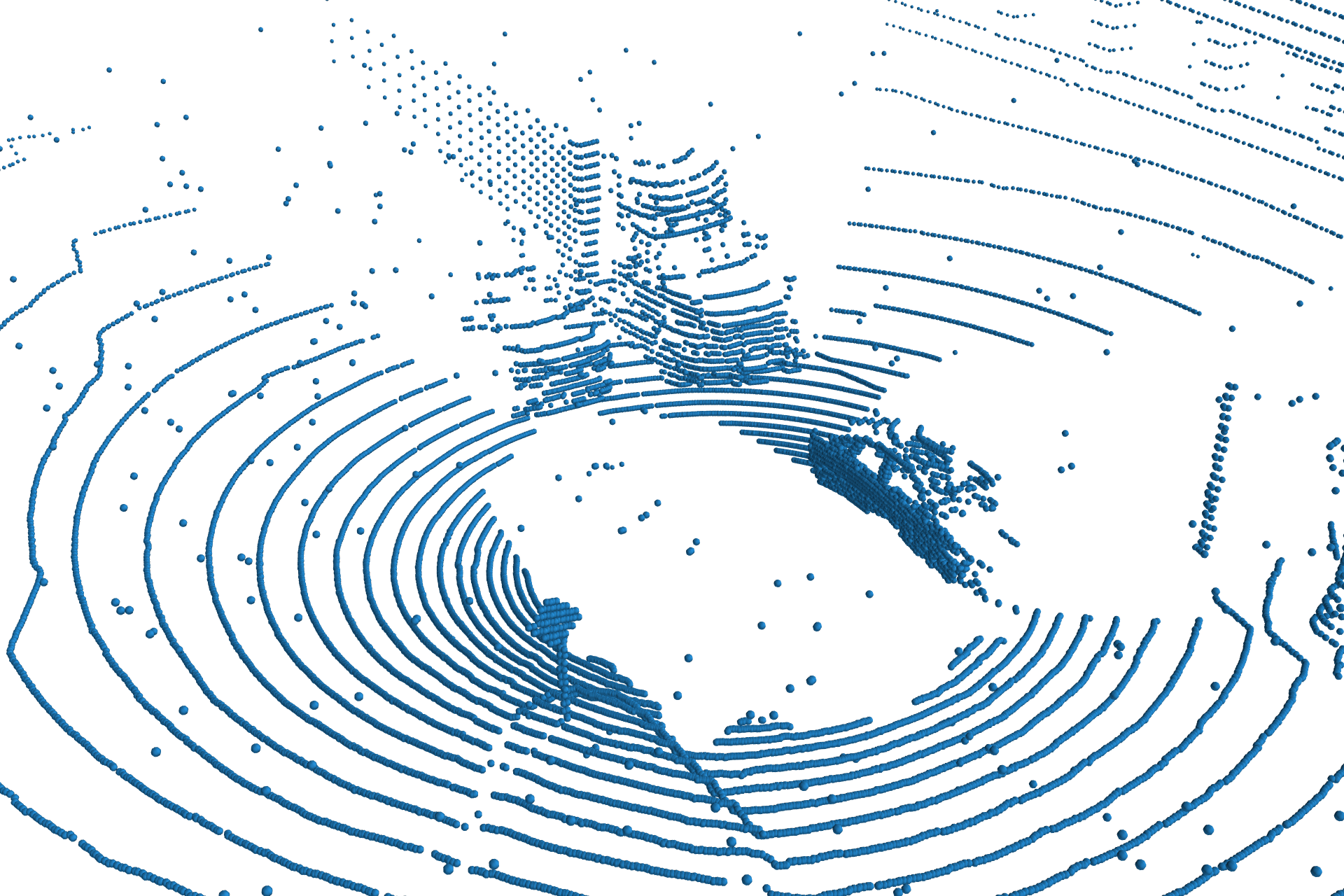}
    \caption{WeatherNet~\cite{heinzler2020cnn}}
    \label{fig_result_32:weathernet1_d}
    \end{subfigure}\hfill
    \begin{subfigure}{0.32\linewidth}
    \includegraphics[width=\linewidth]{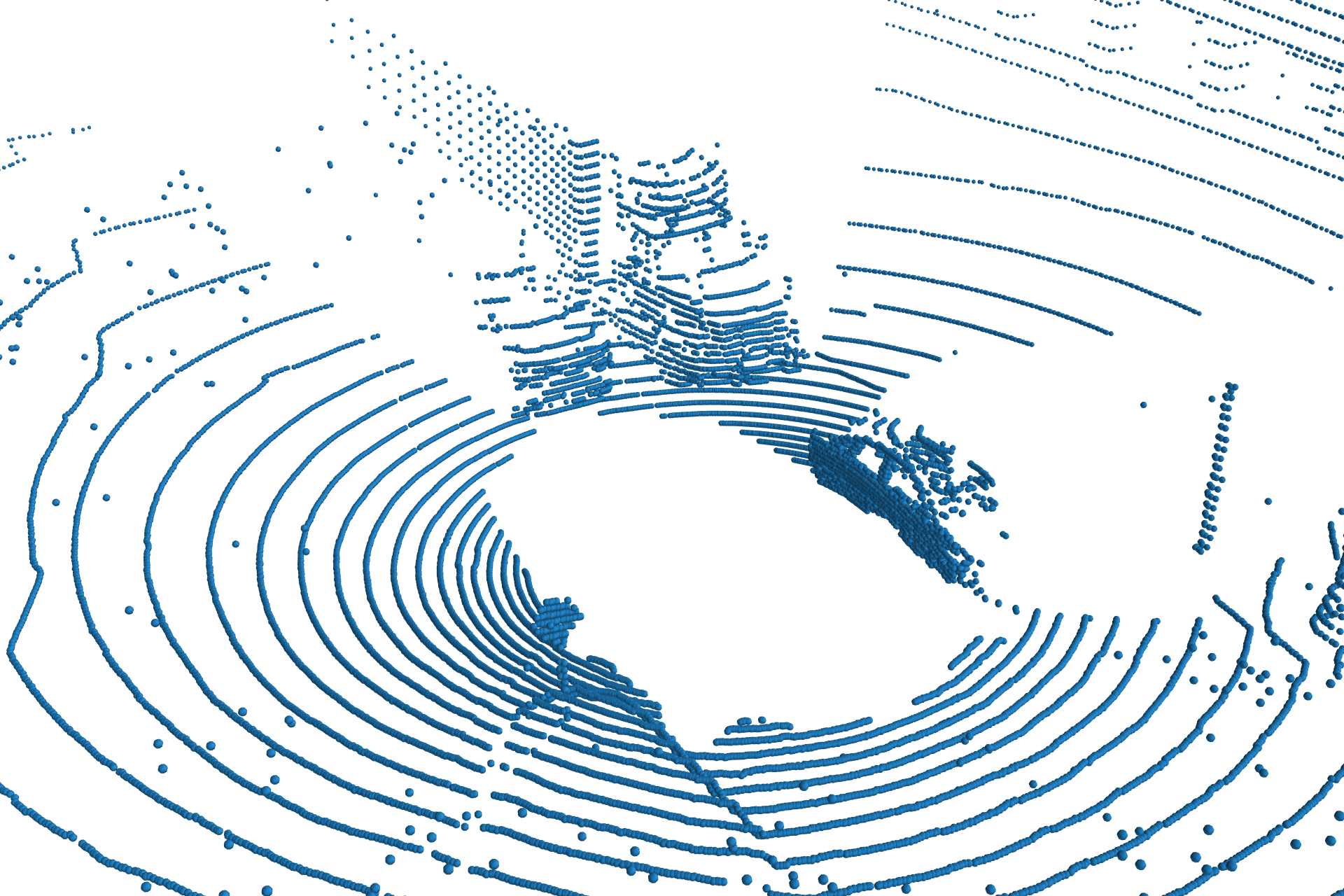}
    \caption{Ours}
    \label{fig_result_32:ours1_d}
    \end{subfigure}\hfill
    
    \caption{Qualitative comparisons on the synthesized snow noise data. The first row shows all points with their prediction results~(\textcolor{myRed}{red}: true positive, \textcolor{myGreen}{green}: false positive, \textcolor{myGray}{gray}: true negative, \textcolor{myYellow}{yellow}: false negative). Classification as noise corresponds to \textit{positive}. The second row shows de-snowed point clouds. DROR misclassifies many sparsely distributed points at the side facade of the truck.}
    \label{fig_result_32}
\end{figure*}

\begin{figure}[t!]
    \centering
    \begin{subfigure}{0.49\linewidth}
    \includegraphics[width=0.49\linewidth]{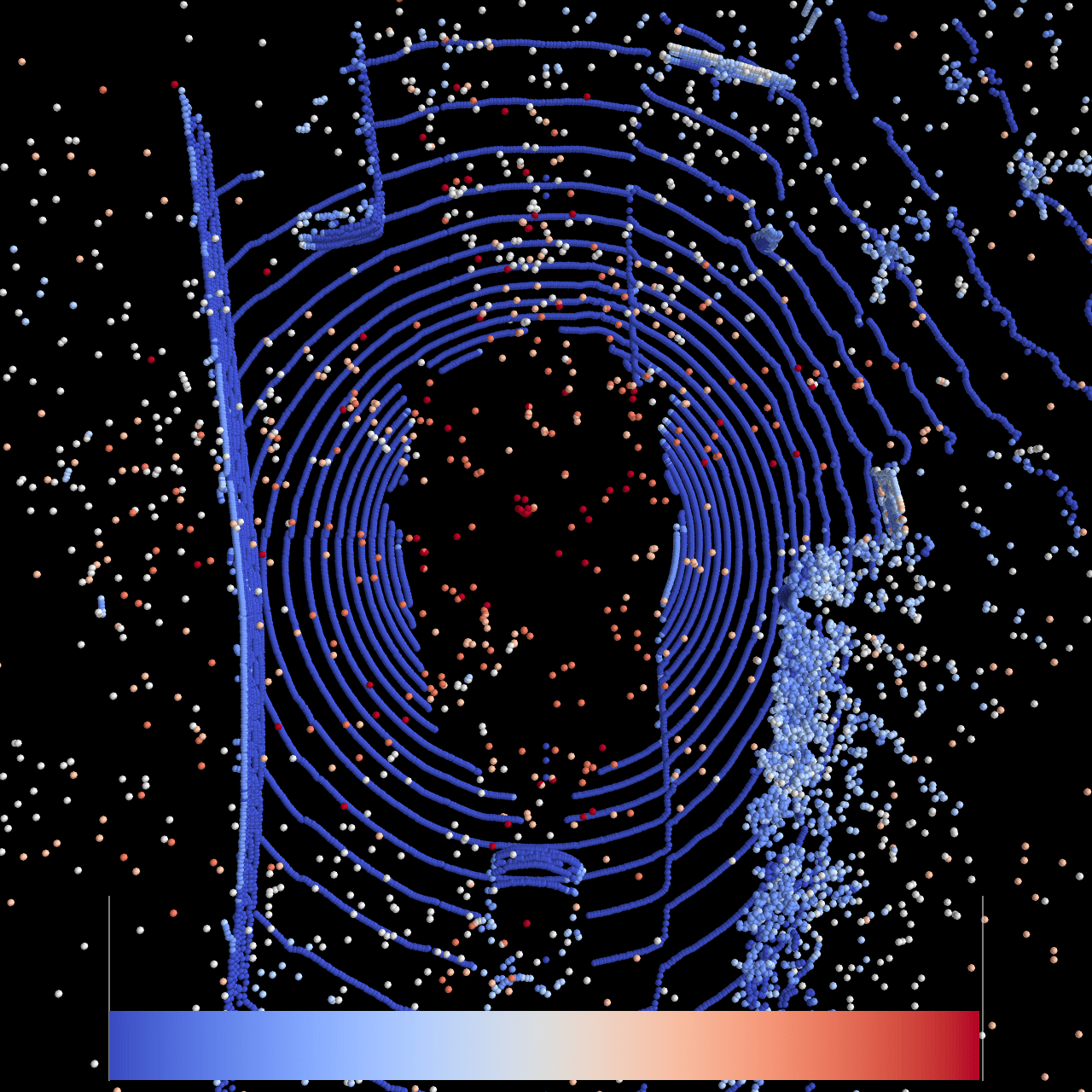}
    \hfill
    \includegraphics[width=0.49\linewidth]{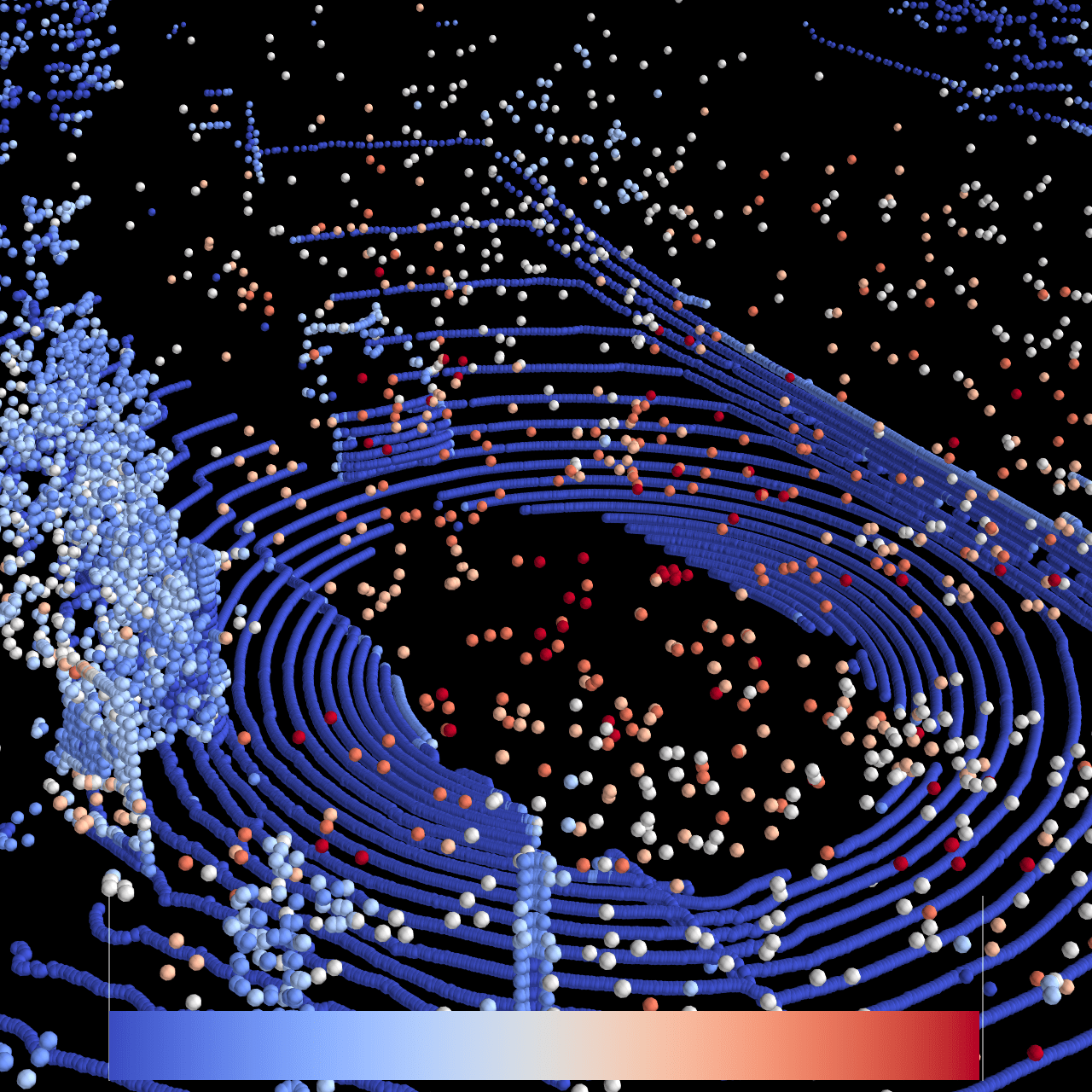}
    \caption{Reconstruction difficulty}
    \label{fig_rd_vis:rd}
    \end{subfigure}\hfill
    \begin{subfigure}{0.49\linewidth}
    \includegraphics[width=0.49\linewidth]{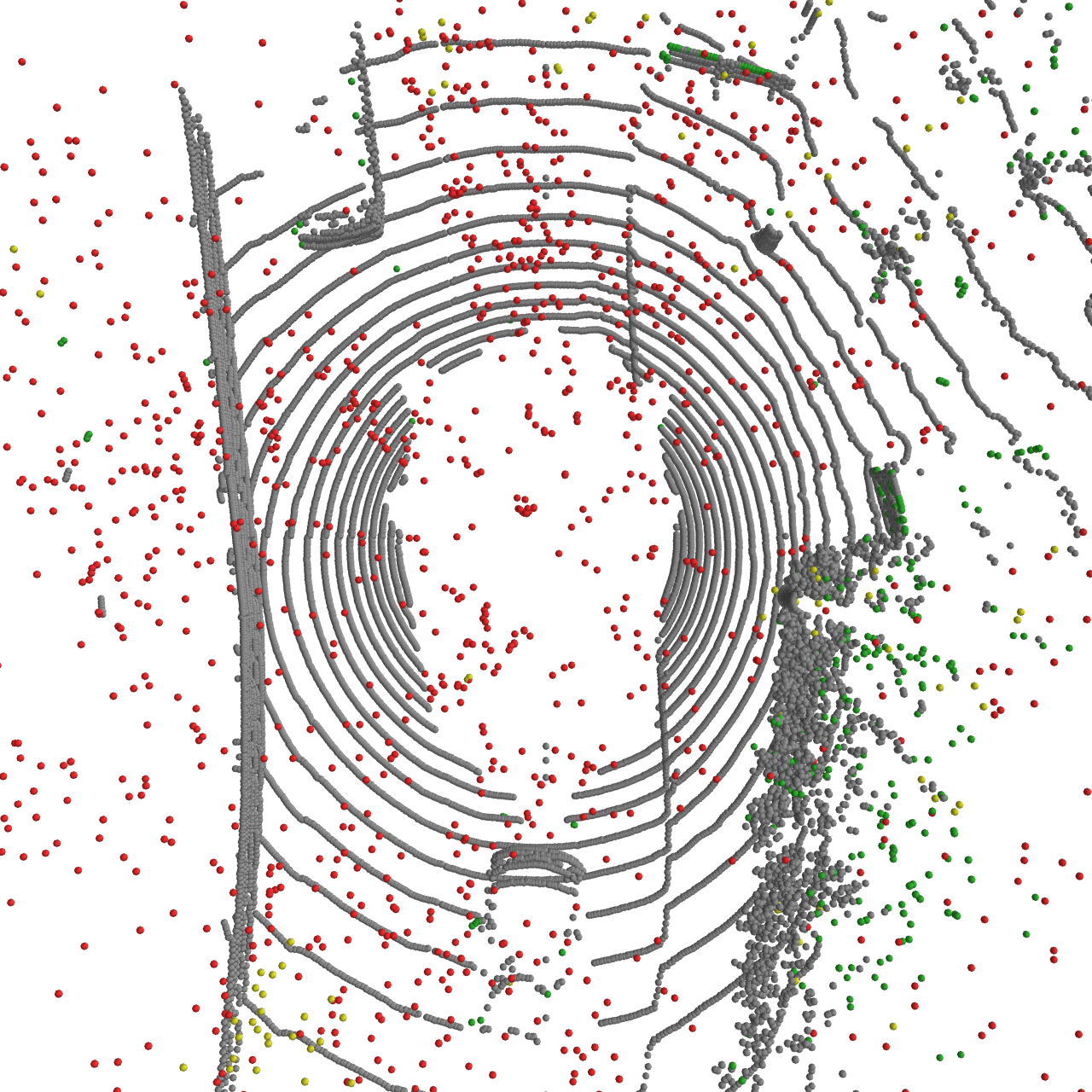}
    \hfill
    \includegraphics[width=0.49\linewidth]{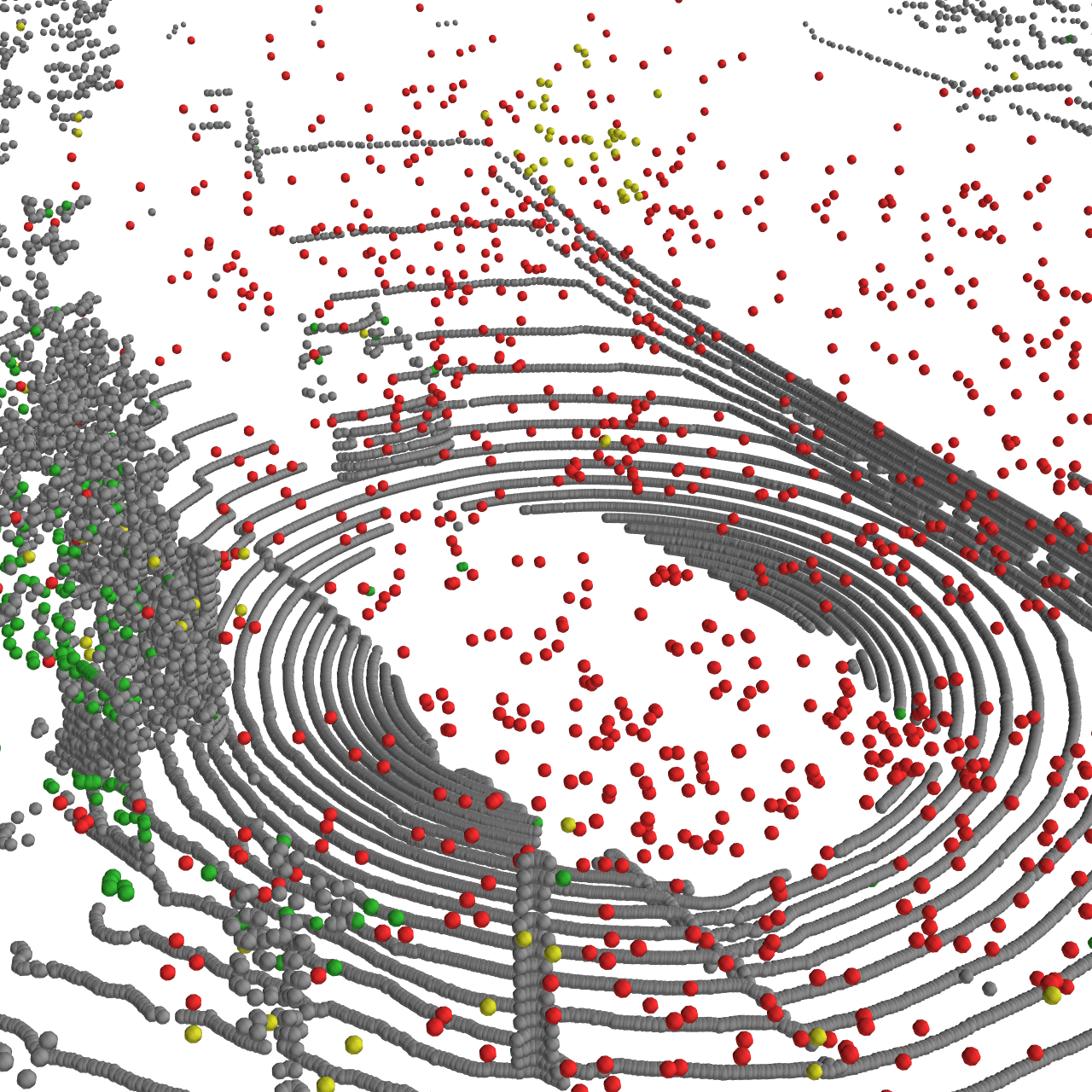}
    \caption{De-snowing result}
    \label{fig_rd_vis:de}
    \end{subfigure}\hfill
    \caption{Estimated reconstruction difficulty and its corresponding de-snowing result. In (a), the closer to the red, the higher reconstruction difficulty. In (b), the color of each points follows~\Fref{fig_result_32}.}
    \label{fig_rd_vis}
\end{figure}

\subsubsection{Quantitative Comparisons}
In~\Tref{table_noise}, our proposed methods are compared with previous LiDAR de-noising approaches: ROR~\cite{rusu20113d}, DROR~\cite{charron2018noising}, and WeatherNet~\cite{heinzler2020cnn}. \Tref{table_noise} shows that our proposed method yields a significantly higher IoU than the state-of-the-art label-free method, DROR. Notably, our approach achieves a comparable IoU to the supervised method without using any labeled data. Applying multi-hypothesis learning in~\Eref{eq_loss_mhl} improved the baseline method in all metrics. For the case where only $1\%$ of labeled data provided, the semi-supervised extension yields significantly higher performance than the supervised method, WeatherNet. Even when $100\%$ labeled data provided, our method performs better than the supervised method without using any additional unlabeled data, which shows better exploitation of the same given data. \textit{Ours~(Semi-sup)} refers to the semi-supervised extension with \textit{Smooth transfer} in~\Fref{fig_weight}, which is described in Section~\ref{sec_semi}.

\begin{table}[t] \centering
    \centering
    \caption{Analyses on the performance changes according to the noise level.}
    \newcolumntype{Y}{>{\centering\arraybackslash}m{1.6cm}}
    \newcolumntype{A}{>{\centering\arraybackslash}m{1.8cm}}
    \newcolumntype{Z}{>{\centering\arraybackslash}m{2.4cm}}
    \begin{tabular}{Z|A|Y Y Y Y}
        \toprule
        \multirow{2}{*}{Method} & \multirow{2}{*}{Metric} & \multicolumn{4}{c}{Noise Level}  \\ \cline{3-6} 
                                &                         & Light & Medium & Heavy & Extreme \\ \hline
        DROR~\cite{charron2018noising}                    & IoU                     & 12.76 & 22.57  & 32.99 & 52.59   \\
        WeatherNet~\cite{heinzler2020cnn}              & IoU                     & 82.29     & 83.57      & 83.57     & 84.60       \\
        Ours(Semi-sup)          & IoU                     & 82.35     & 83.87      & 83.72     & 84.83       \\
        Ours                    & IoU                     & 60.81 & 71.48  & 79.37 & 85.69   \\
        Ours                    & Precision               & 64.35 & 77.81  & 85.48 & 85.69   \\
        Ours                    & Recall                  & 91.70 & 89.78  & 91.74 & 92.42  \\
        \bottomrule
     \end{tabular}
    \label{table_noise_level}
\end{table}

\begin{figure}[t!]
    \centering
    \begin{minipage}[l]{0.47\linewidth} 
    \begin{subfigure}{0.49\linewidth}
    \includegraphics[width=\linewidth]{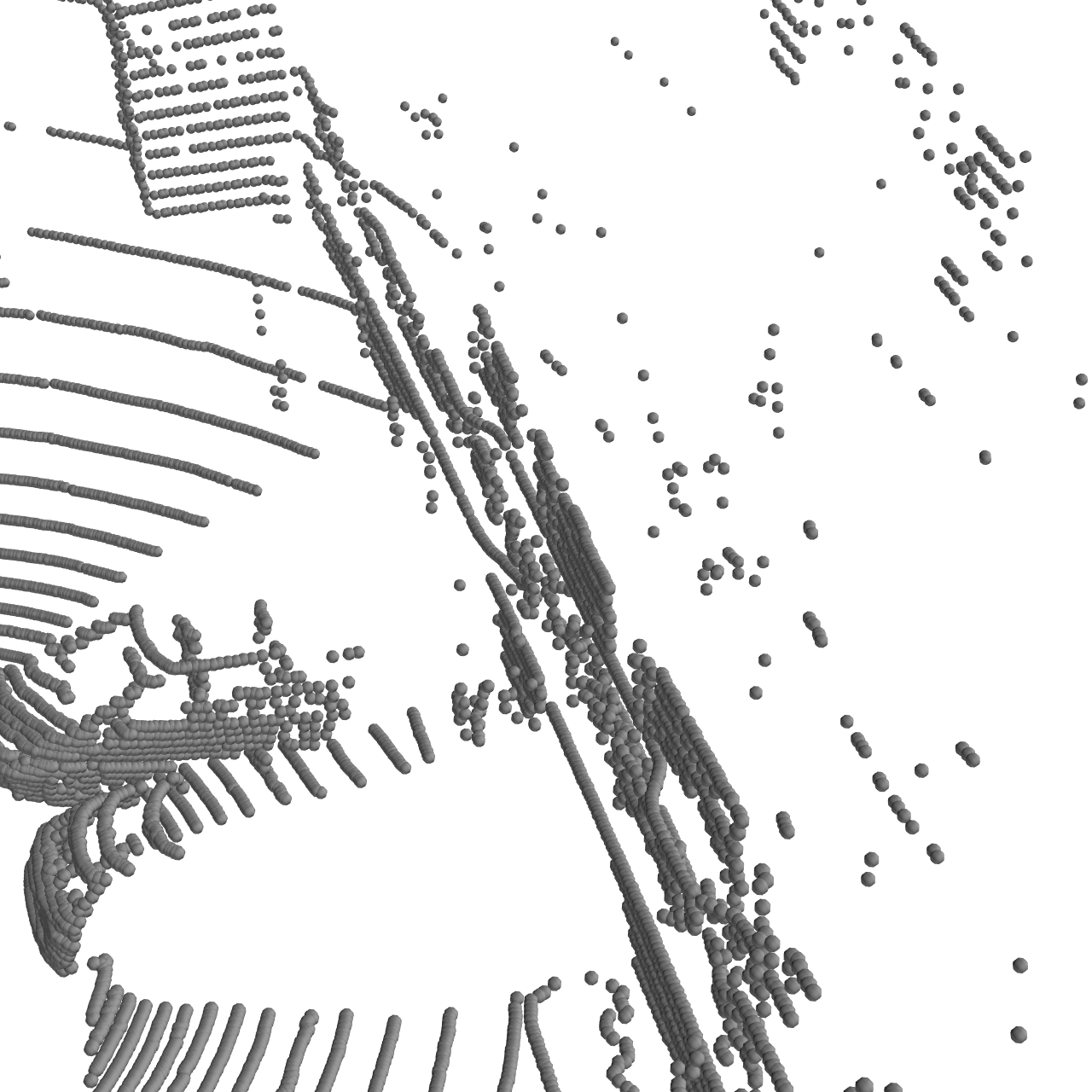}
    \caption{Clean}
    \label{fig_clean:clean}
    \end{subfigure}\hfill
    \begin{subfigure}{0.49\linewidth}
    \includegraphics[width=\linewidth]{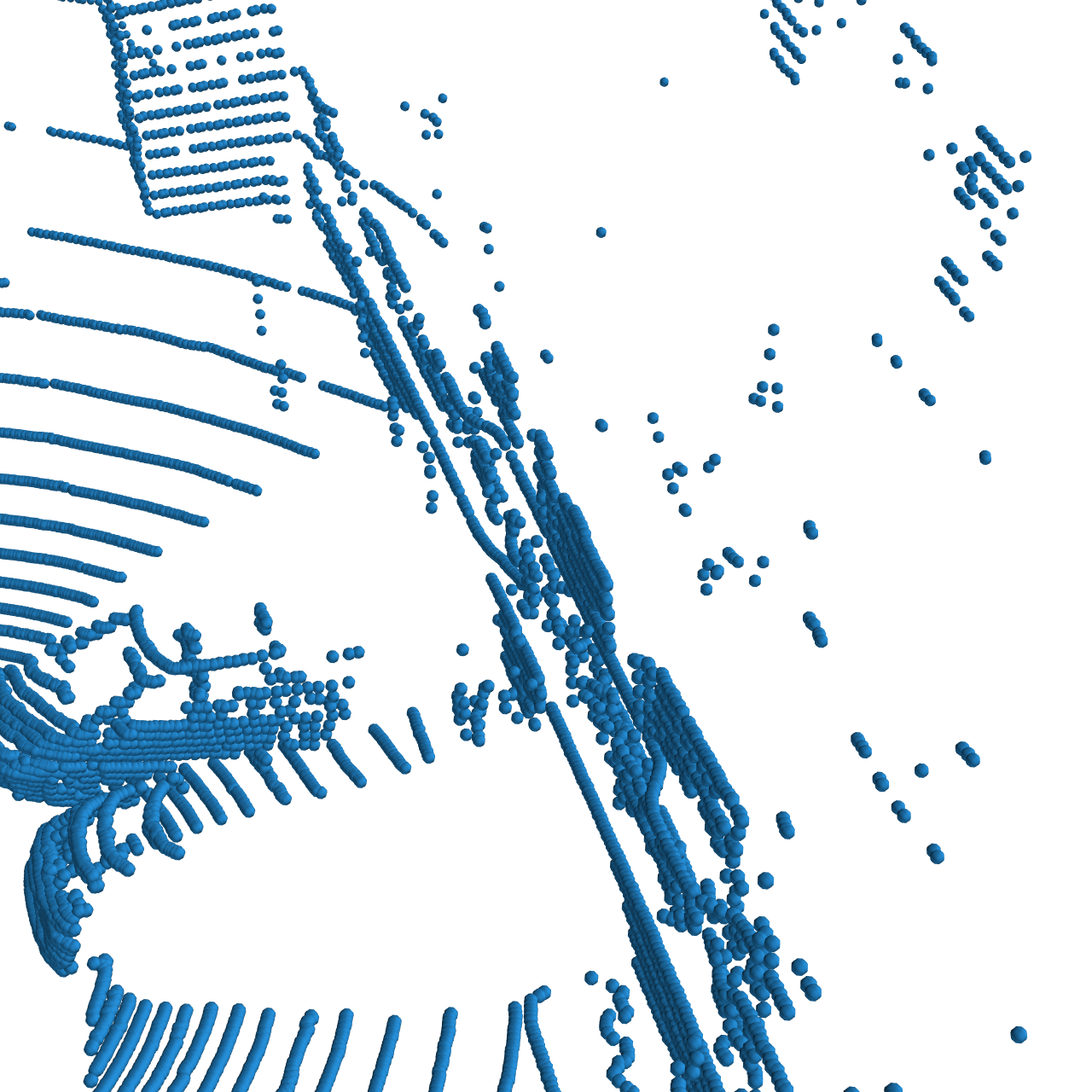}
    \caption{De-snowed}
    \label{fig_clean:denoised}
    \end{subfigure}
    \caption{A de-snowing result of only-clean points scene. Some of floating clean points are similar to noises and eliminated by our method.}
    \label{fig_clean}
    \end{minipage}
    \hfill
    \begin{minipage}[r]{0.47\linewidth} 
    \begin{subfigure}{0.49\linewidth}
    \includegraphics[width=\linewidth]{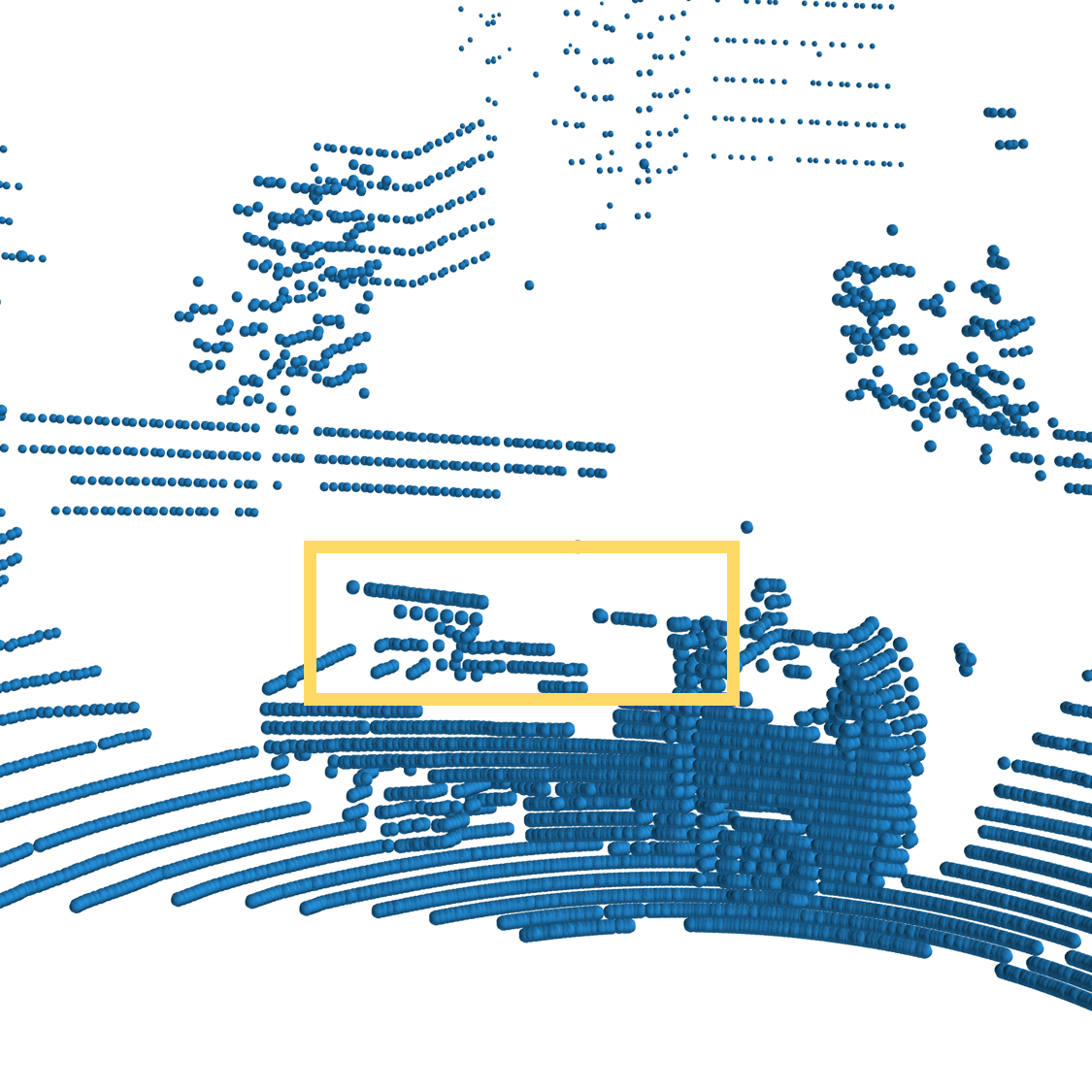}
    \caption{w/o MHL}
    \label{fig_mhl:wo}
    \end{subfigure}\hfill
    \begin{subfigure}{0.49\linewidth}
    \includegraphics[width=\linewidth]{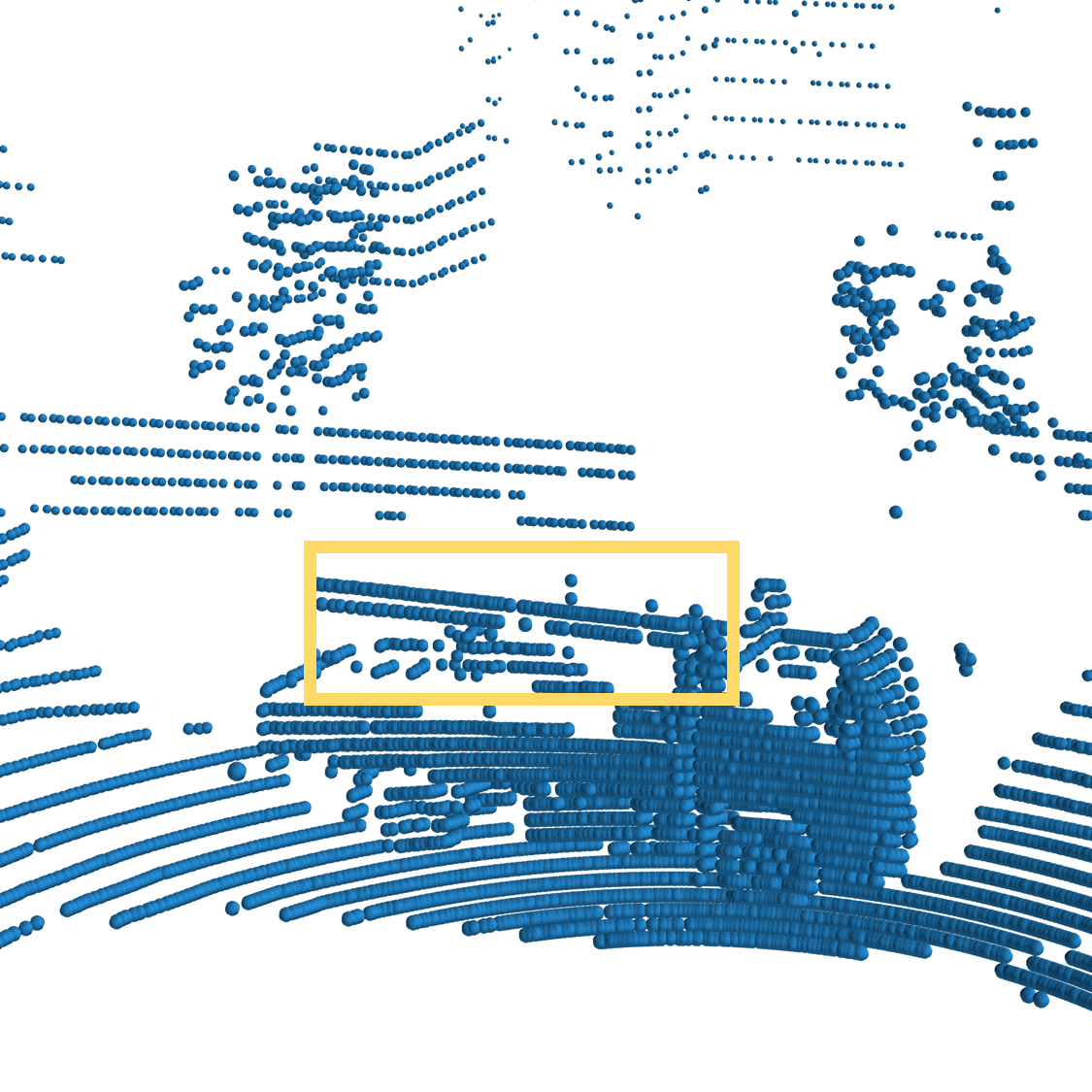}
    \caption{w/ MHL}
    \label{fig_mhl:w}
    \end{subfigure}
    \caption{Comparison between the single hypothesis model and the multi hypotheses model.}
    \label{fig_mhl}
    \end{minipage}
\end{figure}

\begin{comment}
\begin{figure}[t!]
\centering
\begin{tikzpicture}
\begin{axis}[
    x label style={at={(axis description cs:0.5,0.05)}},
    y label style={at={(axis description cs:0.05,0.5)}},
    xlabel={Threshold},
    ylabel={IoU},
    xmin=1.5, xmax=3.5,
    ymin=0.4, ymax=0.8,
    xtick={1.5,2.0,2.5,3.0,3.5},
    ytick={0.2,0.4,0.6,0.8},
    xmajorgrids=true,
    ymajorgrids=true,
    grid style=dashed,
    width=8.8cm,
    height=3.0cm,
]

\addplot[
    color=blue,
    style=very thick
    ]
    coordinates {
    (1.5,0.510011461167)(1.6,0.531519470235)(1.7,0.559113146999)(1.8,0.577512177576)(1.9,0.601623183661)(2.0,0.620765618025)(2.1,0.643871051492)(2.2,0.662540577352)(2.3,0.681342339872)(2.4,0.701895453118)(2.5,0.71404559619)(2.6,0.726731134112)(2.7,0.737815005524)(2.8,0.739604955316)(2.9,0.739553851971)(3.0,0.728902905897)(3.1,0.703323993628)(3.2,0.66917531867)(3.3,0.624751142166)(3.4,0.575337607466)(3.5,0.522391047828)
    };
    
\end{axis}
\end{tikzpicture}
\caption{IoU for different thresholds in validation set.}
\label{fig_uncertainty_threshold}
\end{figure}
\end{comment}

\Tref{table_noise_level} shows an analysis on the performance changes of our method according to the noise level. The noise level is decided by following Canadian Adverse Driving Condition Dataset~\cite{pitropov2020canadian}. While WeatherNet and our semi-supervised model generate relatively consistent performances when the noise level changes, all of the label-free methods including ours have lower IoU as the noise level decreases. Since unsupervised methods do not use direct supervision from point-wise labeled data, noise-like points among clean points can be misclassified as false positives. \Fref{fig_clean} depicts the case where noise-like clean points exist. Although only clean points are displayed, clean points floating in the air look very similar to noise points, and those are removed in~\Fref{fig_clean:denoised}.

In~\Tref{table_semi}, we evaluate three weighting functions proposed for our semi-supervised method in~\Fref{fig_weight}. All of the weighting functions have better performances than the supervised method when limited data are given. Compared to the supervised method, \textit{Ramp up/down}, which is widely used in semi-supervised methods, shows higher IoU when~$1\%$ and $10\%$ labels are given but lower when sufficient labels are given, $100\%$. \textit{Pretrain} yields better IoU than the supervised method even when $100\%$ of labels are given. \textit{Smooth transfer} shows the highest performance when $1\%$ data are labeled and higher than the supervised method even when $100\%$ of data are available for both of methods.

\begin{comment}
\begin{table}[t] \centering
\renewcommand{\arraystretch}{1.2} 
    \centering
    \caption{Recall of clean points~(higher recall means lower number of mis-classified valid points.)}
    \newcolumntype{Y}{>{\centering\arraybackslash}X}
    \newcolumntype{Z}{>{\centering\arraybackslash}m{1.7cm}}
    \begin{tabularx}{\linewidth}{Y|Y Y Z Y}
        \toprule
        Method & ROR~\cite{rusu20113d} & DROR~\cite{charron2018noising} & WeatherNet~\cite{heinzler2020cnn}  & Ours \\
        \midrule
        Recall & 40.15 & 73.31 & 99.72 & 96.68 \\
        \bottomrule
     \end{tabularx}
    \label{table_clean}
\end{table}
\end{comment}

\begin{table*}[t] \centering
    \centering
    \newcolumntype{D}{ >{\centering\arraybackslash} m{3.0cm} }
    \newcolumntype{E}{ >{\centering\arraybackslash} m{2.7cm} }
    \caption{De-snowing performances of the supervised method and our semi-supervised extension when limited labeled data are provided.}
    \begin{tabular}{D|E E E}
        \toprule
        Method & IoU (1\% labels) & IoU (10\% labels) & IoU (100\% labels)\\ \midrule
        WeatherNet~\cite{heinzler2020cnn} & 41.40 & 78.75 & 84.04\\
        Ramp up/down & 79.39 & 81.73 & 82.57\\
        Pretrain & 62.46 & 82.87 & 84.86\\
        Smooth transfer & 82.44 & 83.70 & 84.24\\
        \bottomrule
    \end{tabular}
    \label{table_semi}
\end{table*}

\begin{table*}[t] \centering
    \centering
    \newcolumntype{F}{ >{\centering\arraybackslash} m{2.8cm} }
    \newcolumntype{G}{ >{\centering\arraybackslash} m{2.0cm} }
    \caption{Ablation experiments of the proposed self-supervised method.}
    \begin{tabular}{F|G G G G}
        \toprule
        \# Hypotheses & $1$ & $2$ & $3$ & $4$\\ \midrule
        IoU & $65.75$ & $76.17$ & $79.62$ & $76.52$\\ \toprule
        Blank Ratio & $10\%$ & $30\%$ & $50\%$ & $70\%$\\ \midrule
        IoU & $76.64$ & $78.93$ & $79.62$ & $76.57$\\ \toprule
        Shifting Param. & min & $10^{th}$ & $20^{th}$ & $30^{th}$\\ \midrule
        IoU & $72.63$ & $79.36$ & $79.62$ & $79.22$\\
        \bottomrule
    \end{tabular}
    \label{table_ablation}
\end{table*}

\subsubsection{Ablation Studies} \label{para_study}
In~\Tref{table_ablation}, we investigate performance changes of our self-supervised method according to variations in configurations. First, multi hypotheses learning yields significantly better performance than our baseline method that has a single hypothesis. Among the different number of hypotheses, the model predicting three hypotheses achieves the highest IoU. Second, we analyze the effects of the blank ratio for training. Our method yields the highest performance when $50\%$ of points are blanked. Third, we look into the impact of different shifting parameters in the post-processing step. The experiment with the minimum RD-Net output value for each depth interval as the shifting parameter shows a slightly lower result while other settings achieve similar performance.

\subsubsection{Qualitative Comparisons}
\Fref{fig_result_32} demonstrates a qualitative analysis of our method with DROR and WeatherNet. First, DROR misclassifies clean points if they are sparsely distributed. Assuming the same distance, as an angle between a LiDAR ray and its hitting surface gets farther from perpendicular, a point density of the surface gets lower. As depicted in~\Fref{fig_result_32:dror1_d}, it leads to the failure case of DROR, which solely depends on sparsity for detecting noise. For example, in spite of strong semantic consistency of the side facade of the truck, DROR incorrectly removes the points on it based on sparsity. In contrast, our method preserves clean points if they can be reconstructed from neighboring points. Second, WeatherNet has the smallest number of false positives, which are marked as green points in~\Fref{fig_result_32}. It still has remaining noise points around the LiDAR sensor. More results are in the supplementary material.

\Fref{fig_rd_vis} visualizes the reconstruction difficulty estimated by RD-Net and its evaluation. Clean points have low reconstruction difficulty as they have high spatial correlations with neighbors. On the contrary, RD-Net assigns high difficulty to noise points. In~\Fref{fig_rd_vis:rd}, points on trees have relatively high RD-Net output than the points on the road and cars. It shows that RD-Net successfully learns to reflect the reconstruction difficulty of each point. Although points on trees are inferred as more difficult points than other clean points, they still have lower difficulty than noise points and correctly classified as seen in~\Fref{fig_rd_vis:de}.

\Fref{fig_mhl} demonstrates the effects of multi-hypothesis learning. When PR-Net infers a single hypothesis, many points on the upper side of a car are misclassified and removed, as shown in the yellow box in~\Fref{fig_mhl:wo}. This is the case explained in~\Fref{fig_concept_pr_nd}. On the contrary, as seen in~\Fref{fig_mhl:w}, many of those points are preserved when multi-hypothesis outputs are inferred by PR-Net.

\begin{comment}
\begin{table}[t] \centering
\renewcommand{\arraystretch}{1.2} 
    \centering
    \caption{Ablation study on the number of hypotheses.}
    \newcolumntype{Y}{>{\centering\arraybackslash}X}
    \newcolumntype{Z}{>{\centering\arraybackslash}m{2.2cm}}
    \begin{tabularx}{\linewidth}{Z|Y Y Y Y}
        \toprule
        \# of Hypotheses & 1 & 2 & 3 & 4  \\
        \midrule
        IoU & 68.00 & 72.37 & 75.48 & 73.64 \\
        \bottomrule
    \end{tabularx}
    \label{ablation_mhl}
\end{table}

\begin{table}[t] \centering
\renewcommand{\arraystretch}{1.2} 
    \centering
    \caption{Ablation study on the ratio of the target points.}
    \newcolumntype{Y}{>{\centering\arraybackslash}X}
    \begin{tabularx}{\linewidth}{Y|Y Y Y Y Y}
        \toprule
        Ratio & 5\% & 10\% & 15\% & 20\% & 50\% \\
        \midrule
        IoU & 75.24 & 75.48 & 74.52 & 73.92 & 58.31\\
        \bottomrule
    \end{tabularx}
    \label{ablation_blank_ratio}
\end{table}

\begin{table}[t] \centering
\renewcommand{\arraystretch}{1.2} 
    \centering
    \caption{Ablation study on the shifting parameter}
    \newcolumntype{Y}{>{\centering\arraybackslash}X}
    \begin{tabularx}{\linewidth}{Y|Y Y Y Y}
        \toprule
        Percentile & Minimum & $1^{st}$ & $5^{th}$ & $10^{th}$  \\
        \midrule
        IoU & 74.87 & 75.43 & 75.48 & 75.37 \\
        \bottomrule
    \end{tabularx}
    \label{ablation_shifting_threshold}
\end{table}
\end{comment}

\subsection{Qualitative Evaluation on Real-world Data}

We also evaluate our self-supervised method in real snowy weather scenarios captured by our mobility platform, equipped with a Velodyne `VLS-128'. A total of $9,000$ scans were captured in snowy weather. We qualitatively evaluate our method with DROR, which also does not require point-wise labels. Other experimental settings follow~\Sref{result_synthetic}, except for the height of range images which is set to $128$.

\begin{figure}[t!]
    \centering
    \begin{subfigure}{0.48\linewidth}
    \includegraphics[width=0.49\linewidth]{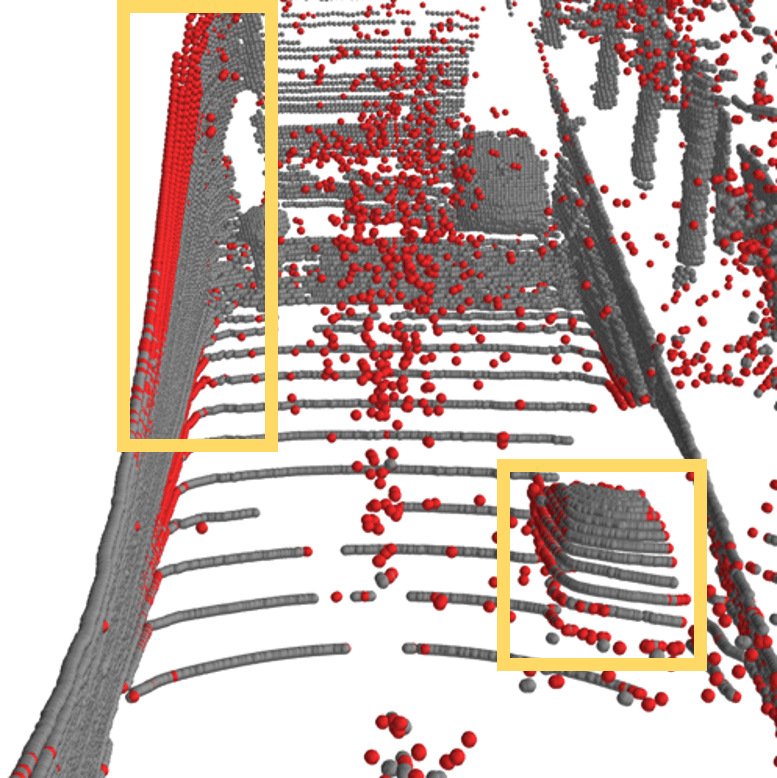}
    \hfill
    \includegraphics[width=0.49\linewidth]{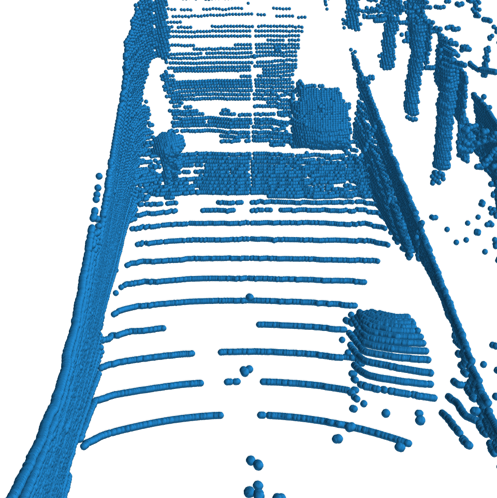}
    \caption{DROR~\cite{charron2018noising}}
    \label{fig_result_128:dror1}
    \end{subfigure}\hfill
    \begin{subfigure}{0.48\linewidth}
    \includegraphics[width=0.49\linewidth]{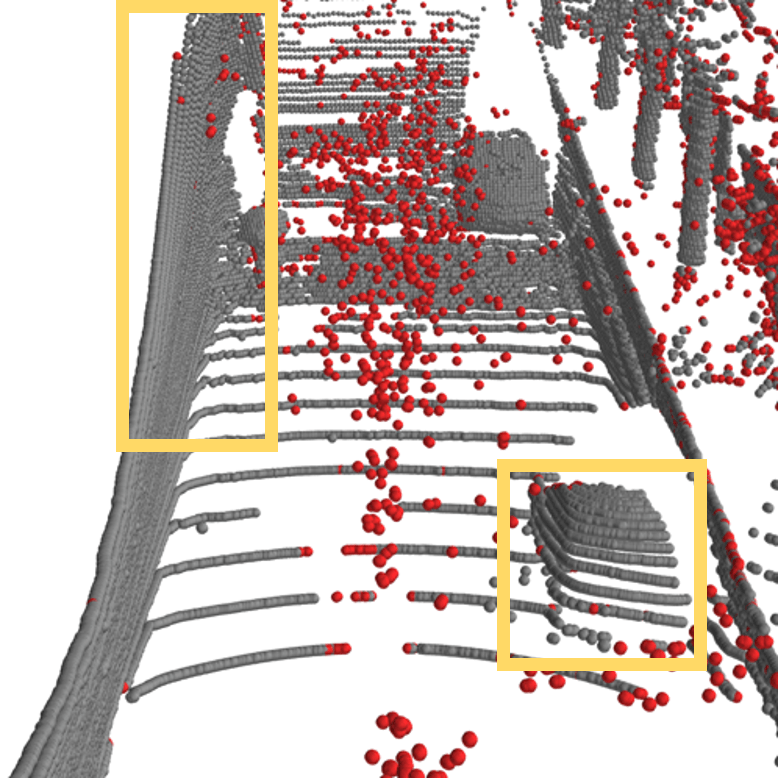}
    \hfill
    \includegraphics[width=0.49\linewidth]{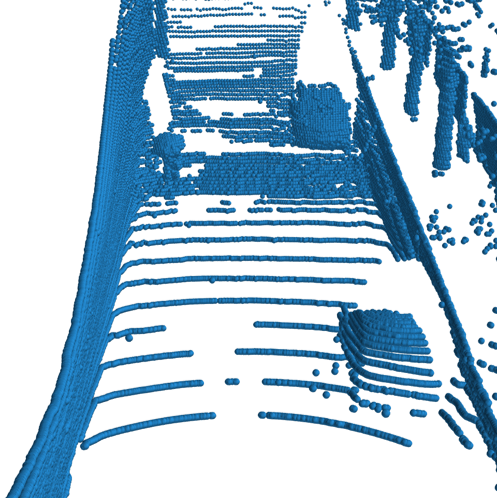}
    \caption{Ours}
    \label{fig_result_128:ours1_d}
    \end{subfigure}\hfill
    \caption{Qualitative comparisons on the real-world snowy weather data~(\textcolor{myRed}{red}: positive, \textcolor{myGray}{gray} and \textcolor{myBlue}{blue}: negative).}
    \label{fig_result_128}
\end{figure}

\Fref{fig_result_128} shows a de-snowing result of DROR and ours. As ‘VLS-128’ LiDAR generates dense point clouds due to its high vertical resolution, the sparsity-based de-snowing method, DROR, generates fewer false positives than in~\Sref{result_synthetic}. However, a number of clean points are still misclassified. For example, as shown in the yellow boxes in~\Fref{fig_result_128:dror1}, clean points on the car and the wall are filtered out when they have low spatial density.
On the other hand, in~\Fref{fig_result_128:ours1_d}, since our method estimates point-wise reconstruction difficulty, clean points on the surface of objects are well preserved. More results are presented in the supplementary material.

\section{Conclusion}
In this work, we proposed a novel self-supervised method for de-snowing LiDAR point clouds in snowy weather conditions. By utilizing the characteristic of noise points that they have low spatial correlations with their neighboring points, our method is designed to detect noise points that are difficult to reconstruct from their neighboring points. Our proposed self-supervised approach outperforms the state-of-the-art label-free methods and yields comparable results to the supervised approach without using any annotation. Furthermore, we present that our self-supervised method can be exploited as a pretext task for the supervised training, which significantly improves the label-efficiency.\\
\textbf{Acknowledgement} This work was supported by the Agency for Defense Development~(ADD) and by the National Research Foundation of Korea~(NRF) grant funded by the Korea government~(MSIT) (No. NRF-2022R1F1A1073505).
\clearpage
\bibliographystyle{splncs04}
\bibliography{references}
\end{document}